%% file: ml-impacts.tex
\documentclass[a4paper,fleqn]{cas-sc}
\makeatletter\let\expandableinput\@@input\makeatother

\usepackage[authoryear,longnamesfirst]{natbib}
\usepackage{units}
\usepackage[group-separator={,}]{siunitx}
\usepackage{subfigure}
\usepackage{wrapfig}
\usepackage[normalem]{ulem} 

\def\tsc#1{\csdef{#1}{\textsc{\lowercase{#1}}\xspace}}
\tsc{WGM}
\tsc{QE}
\long\def\comment#1{}
\newcommand{\todo}[1]{\num{#1}}

\newcommand{\edit}[2]{#2}

\begin{document}
\let\WriteBookmarks\relax
\def\floatpagepagefraction{1}
\def\textpagefraction{.001}

\shorttitle{Using Machine Learning to Reduce Observational
  Biases When Detecting New Impacts on Mars}

\shortauthors{Wagstaff et al.}

\title [mode = title]{Using Machine Learning to Reduce Observational
  Biases When Detecting New Impacts on Mars}



%

\author[1]{Kiri L. Wagstaff}[orcid=0000-0003-4401-5506]
\cormark[1]
\ead{kiri.wagstaff@jpl.nasa.gov}
\ead[url]{https://www.wkiri.com/}
\credit{Concept, Data labeling, Candidate review infrastructure, Writing}

\author[2]{Ingrid J. Daubar}
\credit{Initial impact catalog, Candidate review, HiRISE follow-up
  imaging, Measurements, Data analysis, Writing} 

\author[1]{Gary Doran}
\credit{Globally scalable classifier deployment, Candidate database
  design and implementation, Experimental results, Writing} 

\author[1]{Michael J. Munje}
\credit{Machine learning classifier training, Experimental results}

\author[3]{Valentin T. Bickel}
\credit{Candidate review, Analysis}

\author[2]{Annabelle Gao}
\credit{Candidate review, HiWish HiRISE follow-up requests,
  Measurements, Analysis} 

\author[2]{Joe Pate}
\credit{Candidate review, Analysis, Writing}

\author[2]{Daniel Wexler}
\credit{Candidate review, HiWish HiRISE follow-up requests, Measurements}

\affiliation[1]{organization={Jet Propulsion Laboratory,
  California Institute of Technology},
            addressline={4800 Oak Grove Drive}, 
            city={Pasadena},
            state={CA},
            postcode={91109}, 
            country={USA}}

\affiliation[2]{organization={Brown University},
            addressline={Earth, Environmental, and Planetary Science}, 
            city={Providence},
            state={RI},
            postcode={02912}, 
            country={USA}}

\affiliation[3]{organization={ETH Zurich},
            addressline={Engineering Geology},
            city={Zurich},
            postcode={8092}, 
            country={CH}}

\cortext[1]{Corresponding author}



\begin{abstract}
  The current inventory of recent (fresh) impacts on Mars
  shows a strong bias towards areas of low thermal inertia.  These
  areas are generally visually bright, and impacts create dark scours
  and rays that make them easier to detect.  It is expected that
  impacts occur at a similar rate in areas of higher thermal inertia,
  but those impacts are under-detected.  This study investigates the
  use of a trained machine learning classifier to increase the
  detection of fresh impacts on Mars using CTX data.  This approach
  discovered \todo{69} new fresh impacts that have been confirmed with
  follow-up HiRISE images.  \edit{An examination of}{We found that
    examining} candidates partitioned by thermal inertia (TI)
  values\edit{}{, which is only possible due to the large number of
    machine learning candidates,} helps reduce the observational bias and 
  increase the number of known high-TI impacts.
\end{abstract}


\begin{highlights}
\item Machine learning classifiers can help increase the detection of rare
  fresh impacts on Mars.
\item The most-confident machine learning candidates preserve
  observational biases in the manually identified training examples.
\item Analyzing candidates partitioned by thermal inertia bins reduces
  observational bias.
\end{highlights}


\maketitle

\section{Introduction}
\label{sec:intro}

New impact craters continue to appear on the surface of Mars. The
steadily growing body of observational data acquired by orbital
instruments has enabled the confident identification of more than a
thousand dated recent \edit{}{(``fresh'')} impacts, which
\edit{}{are defined as craters that} are
constrained by images of the same area before and after the impact occurred. 
Newly forming craters were first discovered on Mars by
\cite{malin:crater06}, 
who performed an imaging campaign with the Mars Global
Surveyor Mars Orbiter Camera (MOC) to find twenty new impacts by
manually comparing repeat images of the same areas. These
techniques have continued with the Context Camera (CTX)~\citep{malin:ctx07}
on the Mars Reconnaissance Orbiter for the last $\sim$\num{15} years
\citep{daubar:crater-rate13,daubar:catalog22}. Newly acquired CTX images are
manually searched for distinctive features resembling past discoveries
of new impacts, and then previous imagery from various orbiters is
compared to find the most recent image in which the feature was
absent. If one is found, a follow-up
high-resolution image is requested from the High Resolution Imaging
Science Experiment (HiRISE) on MRO~\citep{mcewen:hirise07}. These new
craters have been rich sources of new information about Mars, for
example studies of subsurface ice
exposures~\citep{byrne:ice09,dundas:ice14,dundas:ice21} 
and mineralogy~\citep{viviano:mineralogy19},
statistics~\citep{daubar:crater-rate13,williams:primaries14,hartmann:craterstats17,daubar:clusters19} 
and morphology~\citep{daubar:morphology14} of current
martian cratering, and investigations of the dust-related albedo
features around the new
craters~\citep{burleigh:avalanche12,daubar:blast16,bart:halos19}. 

\begin{figure}
  \centering
  \includegraphics[width=0.45\textwidth]{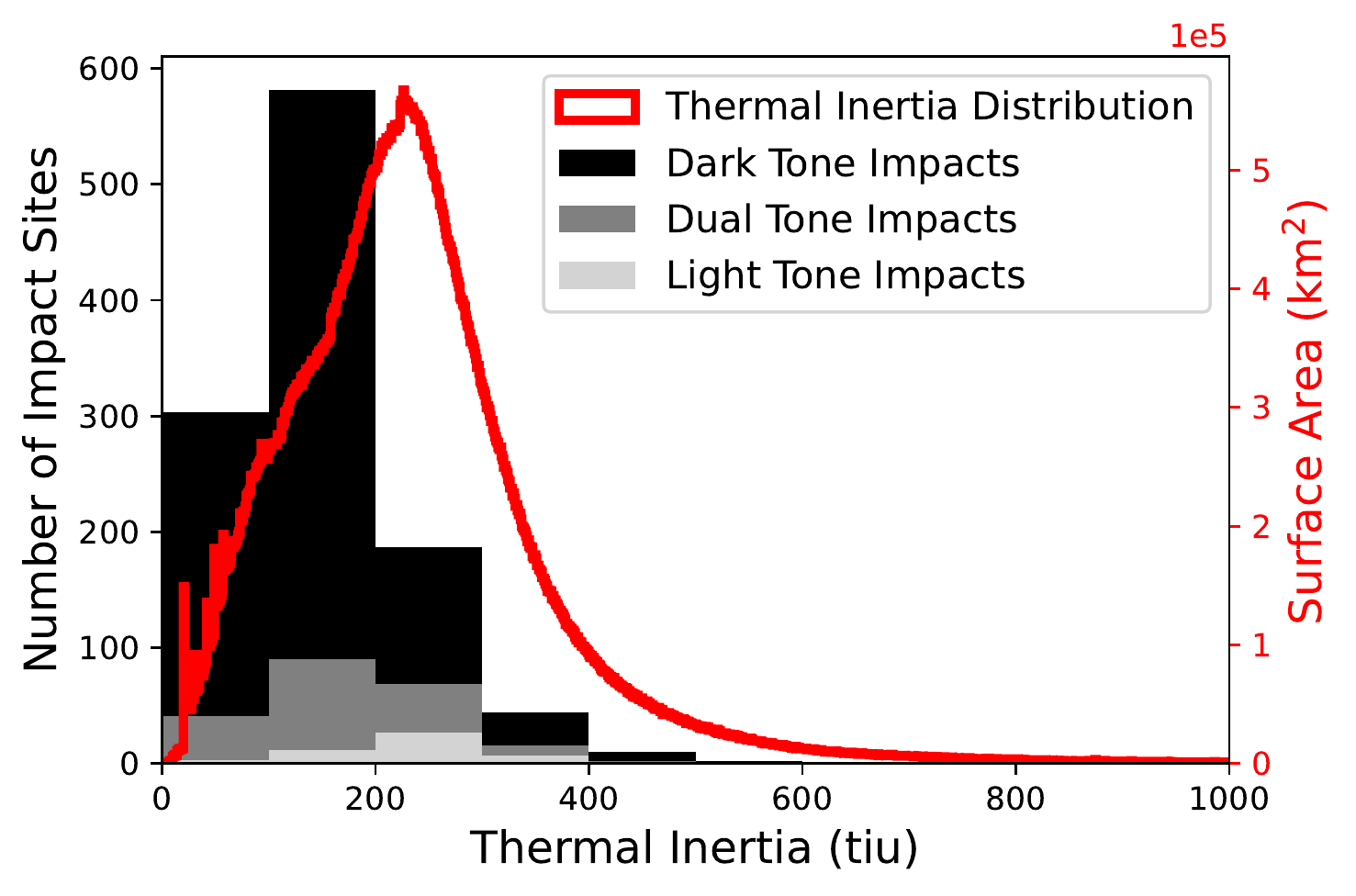}
  \caption{Thermal inertia distribution for fresh impacts \edit{}{from
      \cite{daubar:catalog22}} within -60
    to 60 degrees latitude (histogram) compared with the overall
    thermal inertia distribution for Mars (red curve) over the same
    area, derived from TES and THEMIS thermal inertia maps. The
    stacked histogram shows detections by type. The thermal inertia
    values for fresh impact sites tends lower than that of Mars as a
    whole (peak near 150 tiu vs.~225 tiu).} 
  \label{fig:ti}
\end{figure}

\comment{
\begin{figure}
  \centering
  \includegraphics[width=4in]{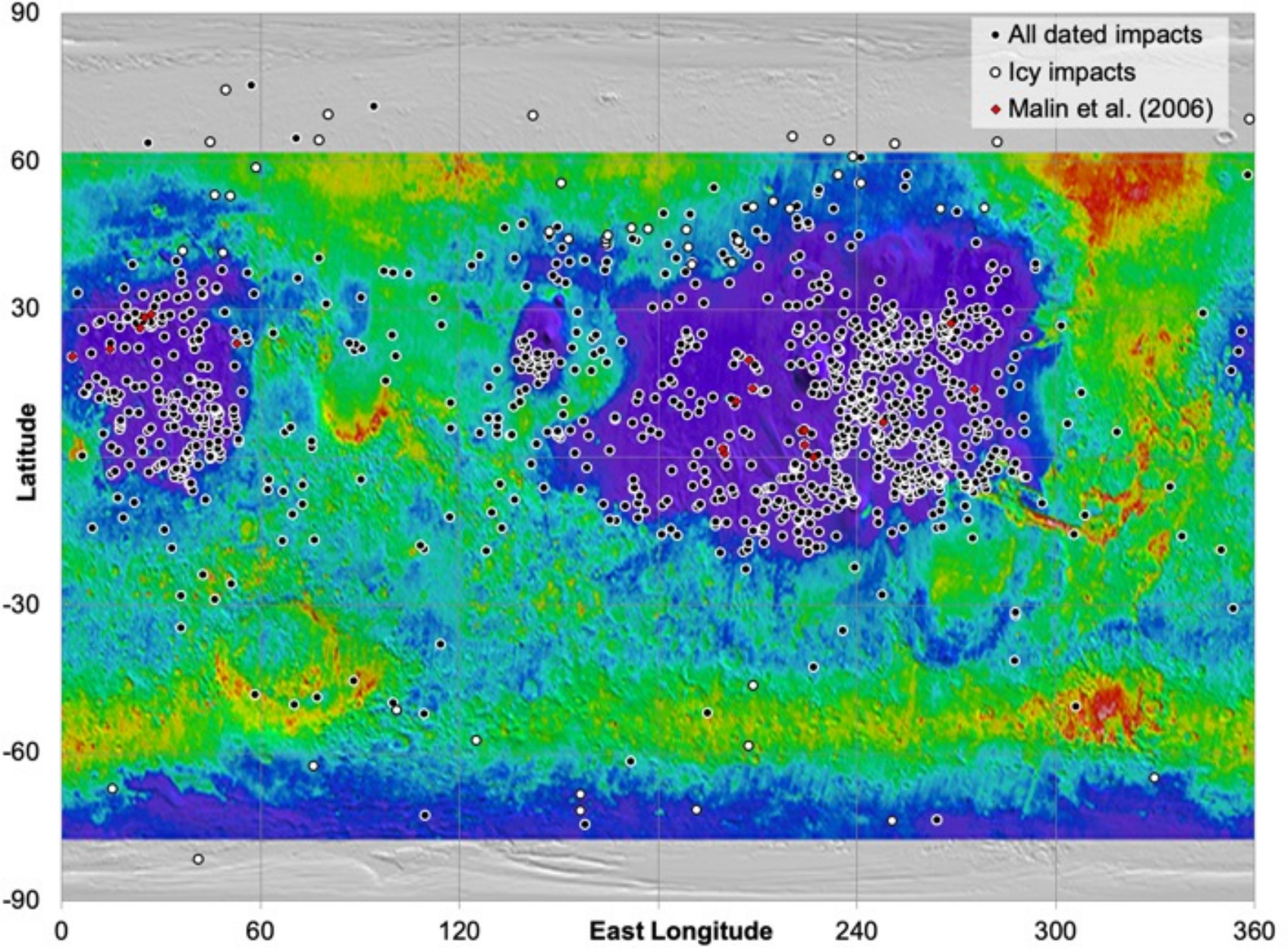}
  \caption{Spatial distribution of fresh impact sites on Mars overlaid
    on the TES thermal inertia basemap~\citep{christensen:tes01}.
    Known impacts are strongly associated with (purple) areas of low
    thermal inertia.} 
  \label{fig:spatial}
\end{figure}
}

One perplexing aspect of the current inventory of known fresh impacts
on Mars~\citep{daubar:catalog22} is that their properties are not
consistent with the global 
properties of Mars.  For example, Figure~\ref{fig:ti} compares the
distribution of thermal inertia values for known fresh impacts versus
the rest of Mars.  \edit{}{Thermal inertia characterizes the ability
  of a material to store and retain heat.  High thermal inertia on Mars
  corresponds to rocky terrain, while low thermal inertia corresponds
  to dusty areas.} In this figure, known impact craters within 60
degrees of the equator are binned by thermal inertia values derived
from the \edit{}{Thermal Emission Imaging System
  (}THEMIS\edit{}{)~\citep{christensen:themis04}} thermal inertia
map~\citep{edwards:themis2011}, using the \edit{}{Thermal Emission
  Spectrometer (}TES\edit{}{)~\citep{christensen:tes01}} thermal inertia
values~\citep{putzig:impacts07} for regions  
where the THEMIS values are unavailable. Stacked bars show a breakdown
of the impacts by appearance in comparison to the overall distribution
of surface area by thermal inertia, similarly derived from the TES and
THEMIS maps. 
The known population of fresh impacts shows a decided
bias towards areas of lower thermal inertia.  These regions are often
visually bright due to sandy or dusty composition, and fresh impacts
often create dark scour marks and rays that stand out to the human
eye.  It is expected that impacts are also occurring at the same rate in
darker areas with higher thermal inertia, but to date they
are under-represented in the inventory of known impacts.
Effectively, rather than cataloguing all fresh impacts on Mars,
existing detection efforts have dominantly identified dark-on-bright
impacts that occur in areas of relatively low thermal inertia.  This
unintentional observational bias limits knowledge of the true
impact cratering rate and affects analyses that rely on crater
activity, such as estimating surface
age~\citep{hartmann:age01,ivanov:age01,hartmann:age05}.  As one
possible 
adverse result, under-counting impacts in high thermal inertia areas
could lead to a cratering chronology that over-estimates surface ages.

The goal of this work is to explore methods for correcting or reducing
this bias to provide a more complete inventory of fresh impacts on
Mars.  There are many reasons why impacts in high thermal inertia (TI)
areas could be overlooked, including their different appearance (often
bright-on-dark instead of dark-on-bright), smaller or missing blast
rays, or human oversight or fatigue.  We hypothesize that training a
machine learning classifier on known examples of low-TI and high-TI
impacts could help improve consistency in detection and coverage of
all impacts.

The contributions of this paper are %
(1) the addition of \todo{69} impacts to the current inventory of
confirmed fresh impacts on Mars, with five more waiting for HiRISE
imaging, %
(2) characterization of key differences in detections found by humans
versus those found by a machine learning algorithm, %
(3) reduction in observational bias achieved by separately analyzing
detections from different thermal inertia settings, and %
(4) discussion of the implications of these findings for the use of
machine learning to aid efforts to catalog\edit{ue}{} features of interest on
Mars and other planetary surfaces.
\edit{}{Section~\ref{sect:methods} describes related work and the
  methods used in this study, Section~\ref{sect:results} presents and
  analyzes the results, and conclusions are summarized in
  Section~\ref{sect:conc}.} 
\edit{}{An important lesson from this study is that the
  highest-confidence candidates found via machine learning exhibited
  the same observational bias inherent in the existing catalog, but the
  large number of candidates found globally enabled an analysis by
  thermal inertia value that reduced this bias.}

\section{Methods: Detecting fresh impacts with machine learning}
\label{sect:methods}

Machine learning methods are increasingly employed to assist in the
analysis of remote sensing data for planetary science and
exploration.  Examples include the classification of lunar soils using
reflectance spectra~\citep{kodikara:lunarclass20},
the classification of terrain types in Mars orbital images to inform
landing site selection~\citep{ono:terrain16,barrett:noah-h22}, 
the classification of rover and orbital images to enable content-based
search of the Planetary Data System image
archives~\citep{wagstaff:deepmars18,wagstaff:deepmars21} or map
features such as volcanic rootless cones and transverse aeolian
ridges~\citep{palafox:cnn17} and rockfalls on
Mars~\citep{bickel:rockfallmars20} and the
Moon~\citep{bickel:rockfallmoon20},  
and the identification of novel features in rover images to accelerate
discovery~\citep{kerner:novel20}.

The current study investigates whether machine learning can also be
effective in detecting fresh impacts in CTX images with the goal of
increasing coverage and reducing observational bias in the catalog.
\edit{}{Several previous investigations have employed machine learning
methods to identify large craters; see~\cite{delatte:mlcraters19}
for a historical overview and discussion of key
issues. \cite{silburt:mooncraters19} trained a U-Net convolutional
neural network (CNN) to detect craters in lunar digital elevation
map images at a resolution of \unit[118]{m/pixel}, and
\cite{jia:mooncraters21} improved the U-Net's performance with an
attention-aware model.  \cite{wilhelm:mars20} labeled over
\num{16000} CTX image cutouts into \num{15} geomorphic surface
classes, including ``crater,'' then compared the classification
performance of six convolutional neural networks.
\cite{lagain:mars21} adapted a CNN that was trained on large craters
in THEMIS data to apply to smaller craters in CTX images and used it
to determine crater populations and infer the age of the ejecta
blankets of ten large craters on Mars.
However, none of these classifiers were trained to detect fresh
impacts, which differ visually from the ``crater'' class in
that fresh impacts often have scour marks and rays and may be so small
that the crater itself is not resolved.}

\edit{}{\subsection{Training and evaluation of the machine learning
    classifier}
\label{sec:classifier}} 

\begin{figure}
  \centering
  \includegraphics[width=1in]{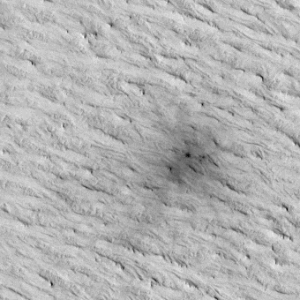}
  \includegraphics[width=1in]{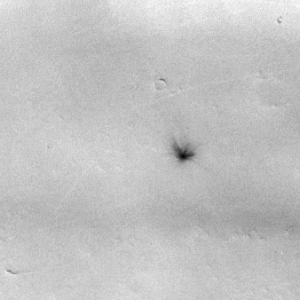}
  \includegraphics[width=1in]{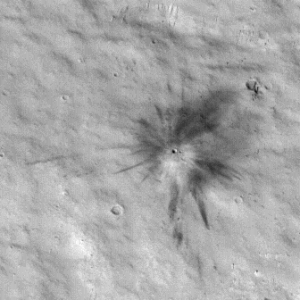}
  \includegraphics[width=1in]{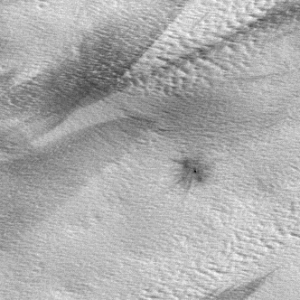}
  \includegraphics[width=1in]{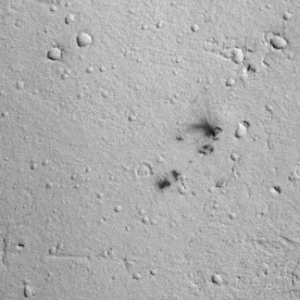}
  \includegraphics[width=1in]{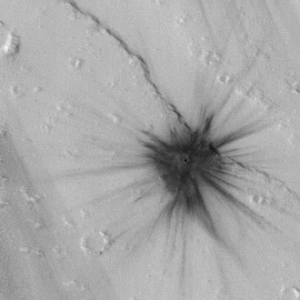}
  \\
  \includegraphics[width=1in]{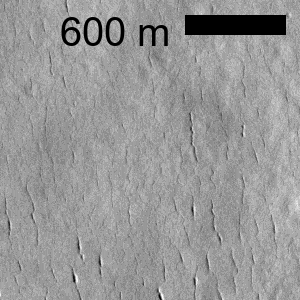}
  \includegraphics[width=1in]{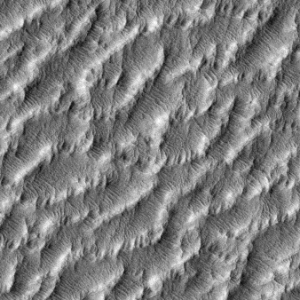}
  \includegraphics[width=1in]{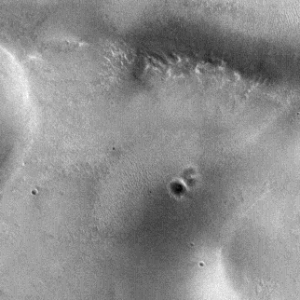}
  \includegraphics[width=1in]{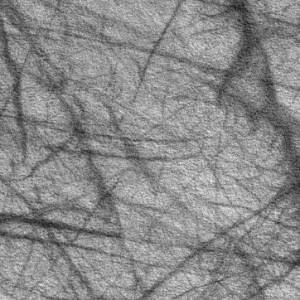}
  \includegraphics[width=1in]{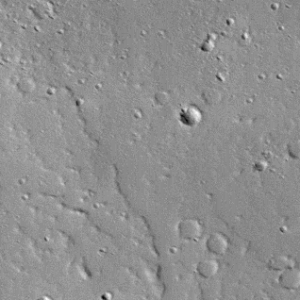}
  \includegraphics[width=1in]{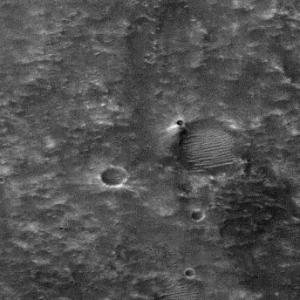}
  \caption{Examples of positive (top row) and negative (bottom row)
    CTX windows used to train the classifier to detect fresh
    impacts. CTX image credits: NASA/JPL/MSSS.}
  \label{fig:ex}
\end{figure}

We trained a convolutional neural network to classify
CTX sub-images (windows) as containing a fresh impact or not.
\edit{We started with a pre-trained
Inception-V3 (Szegedy, Liu, Jia, Sermanet, Reed, Anguelov, Erhan, Vanhoucke and
Rabinovich, 2015) classifier and adapted it by
performing additional training (``fine-tuning'') with examples of
CTX images with and without fresh impacts present.}{}
The windows are small
sections of CTX \edit{}{source} images that span $300 \times 300$
pixels (${\sim}1.8 
\times 1.8$ km).  This size was chosen to allow the detection of fresh
impacts on the scale of a few hundred meters while including some
surrounding context.  We constructed a training set using an existing
catalog of date-constrained fresh impacts that have confirmation from a
HiRISE follow-up
image~\citep{daubar:crater-rate13,daubar:craters20,daubar:catalog22}.
\edit{}{We did not use examples from crater
  databases such as that of~\citep{robbins:craters12} to train the
  fresh impact classifier due to (1) the enormous difference in scale
  (e.g.,~the Robbins database contains craters larger than
  \unit[1]{km} in diameter, but most fresh impacts are only a few
  meters in diameter) and (2) fresh impacts are primarily visible due
  to the surrounding albedo pattern; the craters themselves are not
  generally resolved.  For these reasons, training on larger craters,
  which have a very different visual appearance, would likely be
  detrimental to the search for fresh impacts.} 
Positive examples \edit{are}{were} centered on the location of a known fresh
impact. We manually reviewed each such CTX image to exclude those that
pre-dated the impact's appearance or did not include the impact due to
fading \edit{}{over time} or poor localization.
Negative examples were obtained from a spatially uniform random
sampling of CTX images across the surface of Mars.  In all, we
obtained \num{1856} positive and \num{4973} negative images (total
$n=$ \num{6829}); examples
are shown in Figure~\ref{fig:ex}, and the full data set can be
downloaded at \url{https://zenodo.org/record/5523886}.

We randomly separated \edit{these}{the CTX} images into 90\% training
($n=$ \num{6156}) and 10\% held-out data ($n=$ \num{673}).
We increased the number of training examples by applying standard
image augmentation techniques to \edit{generate additional examples that
capture changes in the images that are expected not to change its 
designation as a positive or negative example}{modify the image
  without changing its class}.  \edit{These
  include}{Each training image was} \edit{flipping each image}{flipped}
horizontally \edit{or}{and} vertically;
rotat\edit{ing}{ed} by \edit{}{a random choice of} 90, 180, or 270
degrees; \edit{}{adjusted by a} minor (random)
\edit{adjustments}{change} to brightness, contrast, 
saturation, and hue; and \edit{applying}{blurred with} a Gaussian
filter with radius 2\edit{ to induce minor blurring}{}.  In
total, we employed \num{36936} images (\num{10080} positive and
\num{26856} negative) images to train the classifier. 

\edit{}{We used the state-of-the-art Inception-V3 image
  classifier~\citep{szegedy:inception15}, which was originally trained
  on \num{1.2} million images from the ImageNet data
  set~\citep{ILSVRC15}.  CNN classifiers do not require the
  pre-specification of relevant features and instead automatically
  identify image properties that correlate with the classes of interest.
  We adapted Inception-V3 to classify fresh impacts in
  CTX images by performing additional training (``fine-tuning'') using
  the positive and negative CTX examples.  The fine-tuning approach is
  generally more effective than training a randomly initialized neural
  network from scratch, since the pre-trained network has already
  learned general image properties from the much larger original data set.} 
\edit{}{Inception-V3 employs \num{42} processing layers that include
  efficiency improvements to reduce the number of weights required and
  therefore the training time.}
%
\edit{}{To fine-tune this model for fresh impact detection, we used
  the PyTorch library~\citep{paszke:pytorch19} with the Adam optimizer
  and cross entropy loss function.  Other parameters include a
  starting learning rate of \num{4e-6} with a decay of \num{0.1} every
  \num{10} steps, batch size of \num{32}, and \num{25} epochs of
  training.}  
\edit{}{The held-out set was used to calibrate the
  classifier's predictions (posterior probability of a fresh impact)
  using bias-corrected temperature 
  scaling}~\citep{alexandari:bcts20}. Calibration compensates for classifier 
over- or under-confidence by adjusting \edit{}{(rescaling)} the output
class probabilities so that they correspond to true occurrence rates
in the held-out data set.
\edit{}{The calculated scaling parameter (temperature) was \num{1.51}
with class bias terms of \num{-0.05} (negative, non-impact) and
\num{0.26} (positive, fresh impact).  The calibration process
reduced the Expected Calibration Error~\citep{naeini:ece15} from
\num{0.017} to \num{0.009} in the posterior probability values.  The
classifier achieved an accuracy of \num{97.5}\% (with precision
\num{0.98}, recall \num{0.93}, and F1-score \num{0.95}) on the
held-out data set.} 

\subsection{Deployment of machine learning classifier to the CTX archive}

We deployed the trained classifier on the entire CTX archive,
excluding images from the Mars Orbit Insertion and Cruise phases of
the mission, as well as a few observations that caused issues because
they were lacked valid geolocation information (e.g.,~calibration images).
The resulting set contained \num{112207} observations 
acquired between September 27, 2006 and December 1, 2019.
We used the
Gattaca supercomputer at JPL to process the observations in parallel
using between \num{500} and \num{750} CPU cores over a period of about
one week. Each observation was split into overlapping square windows
\unit[300]{pixels} in size, sliding \unit[75]{pixels} at a time in each
direction. The classifier was evaluated on each window, for a total of
approximately 2~billion individual classifications. The total computation time
across all cores was roughly \unit[12]{CPU-years}.

After \edit{computing}{generating} classification posterior
probabilities \edit{}{with the classifier} for each 
window, the full set of classifications was ranked in descending order
of the probability that a 
fresh impact was present. Within the CTX archive, there may be multiple
observations of the same location obtained at different times. Therefore, for
each classification in the descending list, we grouped other windows that were
found within \unit[600]{m} of the classified window into a single ``candidate''
for manual inspection. Each window is associated with only a single
candidate to avoid duplication.

The list of candidates was filtered for further analysis.  First,
\edit{}{informed by previous
  studies~\citep{dundas:ice14,daubar:blast16},} we only considered
candidates within the latitude range from \num{-60} to
\unit[60]{degrees}. \edit{}{High-latitude impact craters disappear 
  quickly due to seasonal processes, and there is a lack of reliable
  training examples due to their rarity, seasonally poor illumination
  conditions, and seasonal changes in surface appearance due to ice
  and frost.}
Furthermore, we filtered out high-probability
candidates for which there was not a least one non-detection (an
overlapping window with a classification score below 0.5) within the
candidate set that was acquired earlier.
This excludes candidates for which the CTX data likely does not
contain sufficient temporal coverage of the impact location to constrain its
date of formation (i.e., an image the pre-dates the impact).
After the filters were applied, a list of the
\num{1000} highest-confidence candidates was generated for review.


\subsection{Manual review of machine learning candidates}


\begin{figure}
  \centering
  \fbox{\includegraphics[width=4in]{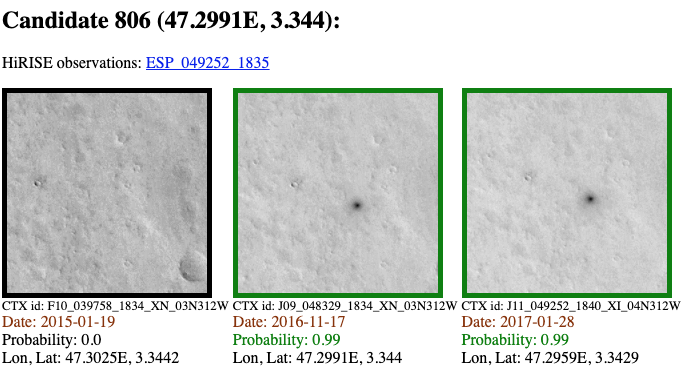}}
  \caption{Information displayed for a fresh impact candidate found by
    the machine learning classifier to enable manual review.
    This candidate consisted of three CTX observations (the images are
    map-projected with north up).  A link
    to any pre-existing HiRISE observations of the same area is
    provided to enable a high-resolution view when available and to
    facilitate identification of previously known fresh impact sites
    that had already been imaged by HiRISE.  Note that since windows
    were independently generated by stepping every 75 pixels through
    different source CTX images, they are not perfectly registered.
    CTX image credits: NASA/JPL/MSSS.} 
  \label{fig:cand}
\end{figure}

We performed a manual review of the top \num{1000} candidates
identified by the classifier.  The candidates were ranked in
descending order of posterior probability as reported by the
classifier\edit{}{; more than one million candidates had probability $\geq
  0.99$}.  To facilitate this process, we generated a web page that 
provided information for each candidate, including all images of the
candidate site (including repeat CTX observations, if available).  We
included only candidates whose first CTX observation was a
non-detection (negative) to allow us to constrain the time of
formation.  Reviewing the complete time series of the location helps
determine whether it is a valid fresh impact.  Figure~\ref{fig:cand}
shows an example candidate that consists of three CTX observations of
the same location. 
Observations in which the classifier predicts a fresh impact
probability of at least \num{0.95} are automatically outlined in
green.  The first observation does not contain an
impact and was given a probability of \num{0.0}, indicating that 
this impact occurred between January 19, 2015 and November 17, 2016.
The review page also includes a link to any overlapping HiRISE
observation(s) that may already exist, enabling the reviewer to easily
check for the presence of the impact at the time the HiRISE
observation was collected.


The review process for each candidate proceeded as follows.
Reviewers first examined the set of detection images for any
newly appearing albedo features.  In some cases, the
classifier mistakenly assigned high probability to features that are
not impacts, such as shadows from surface topography,
dust devil tracks, wind or slope streaks, or other surface changes.  Reviewers
classified any such incorrect detections as ``non-impact.''
Impacts that appeared to be old and could not be date-constrained with
a CTX image of the same location lacking the feature were designed
``old impact.''  Otherwise,  
reviewers checked for the earliest non-detection to constrain
the date of formation.  This enables us to determine with confidence
that the feature is a ``fresh'' impact (occurred within the CTX
operational period).  If no preceding non-detection was available, but
the albedo feature resembled a fresh impact blast zone, 
reviewers classified the candidate as an ``undateable fresh impact.''
Dateable fresh impacts were further divided into ``known fresh
impact'' (already in the catalog) or ``new fresh impact'' (not
previously known).  In some cases, the classifier generated duplicate
candidates from different observation dates that were not grouped due
to localization errors.  Reviewers marked these as ``duplicate.''
Reviewers also
added other notes to aid interpretation where useful.
%
Four reviewers contributed to the assessment of the top \num{1000}
candidates to split up the work.  At the beginning of the review
process, all reviewers examined the same set of roughly \num{50} candidates,
and any disagreements were discussed and resolved to ensure consistent
evaluation of the remaining candidates.
  
For each candidate that was marked as ``new fresh impact'' (not previously
known), we entered HiWish requests
(\url{https://www.uahirise.org/hiwish/}; \cite{chojnacki:hiwish20}) to
obtain a high-resolution view of the same location for confirmation of
the presence of an impact and measurements of crater characteristics. These 
targets were acquired by HiRISE between October 2020 and December
2021, resulting in \unit[0.25]{m/px} images of each site.

\subsection{Measurements of fresh impact features}
\label{sec:measure}

  
To examine the ML-detected new impacts in the context of the
broader human-detected catalog presented by~\cite{daubar:catalog22}, 
the same characteristics were measured for each ML-detected
impact site. Specifically, we examined the follow-up HiRISE images of
promising impact sites to catalog\edit{ue}{} their features,
including whether they contained a single crater or cluster of
craters.  We also characterized the albedo pattern around the
impact site (diffuse halo, linear or arcuate rays, dark-, light- or
dual-toned blast zone in comparison to the surroundings). Crater
diameters were measured following the methodology of~\cite{daubar:crater-rate13,daubar:clusters19,daubar:catalog22},
using the CraterTools add-in for the Java
Mission-planning and Analysis for Remote Sensing,
JMARS~\citep{christensen:jmars09}.  For impacts that consisted of
crater clusters,  
the individual diameters of the craters in the cluster were combined
into an effective diameter that approximates the diameter that
would have occurred had the impactor not fragmented in the
atmosphere.
\edit{}{The effective diameter was calculated as}
\begin{equation}
  \edit{}{D_{\mathrm{ef\/f}} = \sqrt[\leftroot{-2}\uproot{2}3]{\sum_i D_i^3},}
\end{equation}
\edit{}{
where $D_i$ is the diameter of each individual
crater~\citep{malin:crater06,daubar:crater-rate13}.}

Regional characteristics were obtained from
quantitative maps in JMARS using map sampling of the relevant basemap
to calculate the average value within the footprint of the HiRISE
image. This provides a representative value of the region, because the
impacts themselves are orders of magnitude smaller than the basemap
resolutions. Elevations were sampled 
from the MOLA 128 ppd basemap~\citep{smith:mola01}. Dust Cover was
sampled from the TES Dust Cover Index~\citep{ruff:tes02}.
Thermal inertia was sampled from the THEMIS quantitative 100
m per pixel global mosaic \citep{edwards:themis2011}. Where the
THEMIS data were not available, we used the average of day and night
TES thermal inertia~\citep{putzig:impacts07}. 

\begin{figure}
  \centering
  \subfigure[June 21, 2007 (CTX {\scriptsize P08\_004222\_1917\_XI\_11N338W})]
  {\includegraphics[height=1.7in]{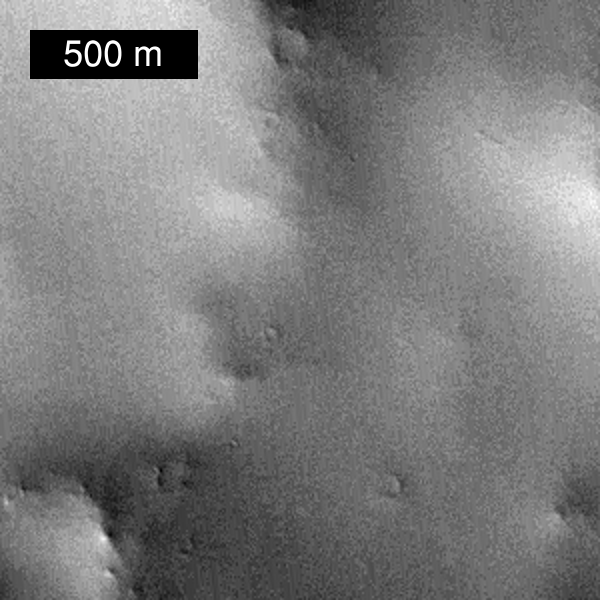}}
  \subfigure[May 13, 2014 (CTX {\scriptsize F02\_036541\_1943\_XI\_14N338W})] 
  {\includegraphics[height=1.7in]{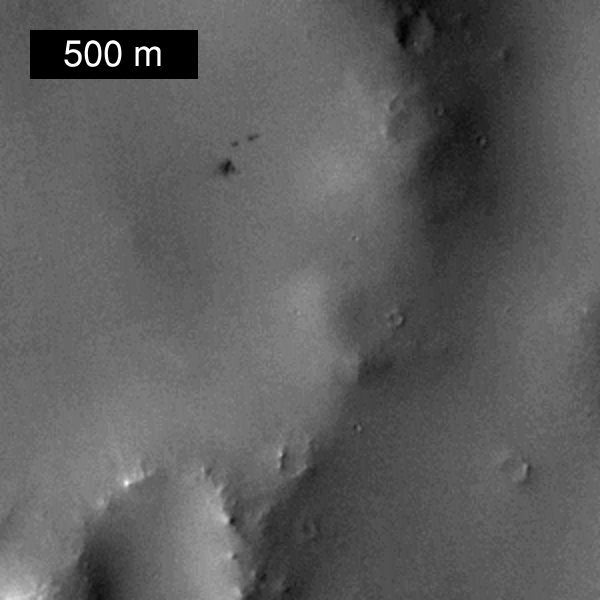}}
  \subfigure[April 2, 2021 (HiRISE ESP\_068824\_1920) at 11.79$^\circ$
  N, 21.48$^\circ$ E]
  {\includegraphics[height=1.7in]{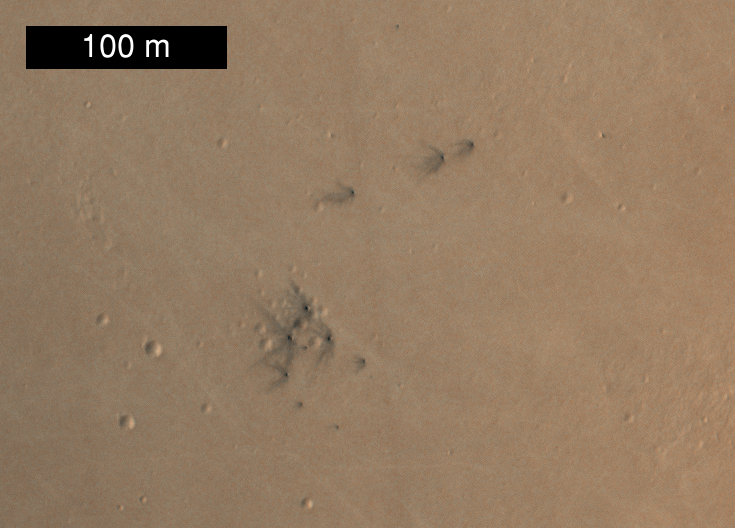}}

  \subfigure[October 24, 2007 (CTX {\scriptsize P12\_005822\_1709\_XI\_09S0980W})]
  {\includegraphics[height=1.7in]{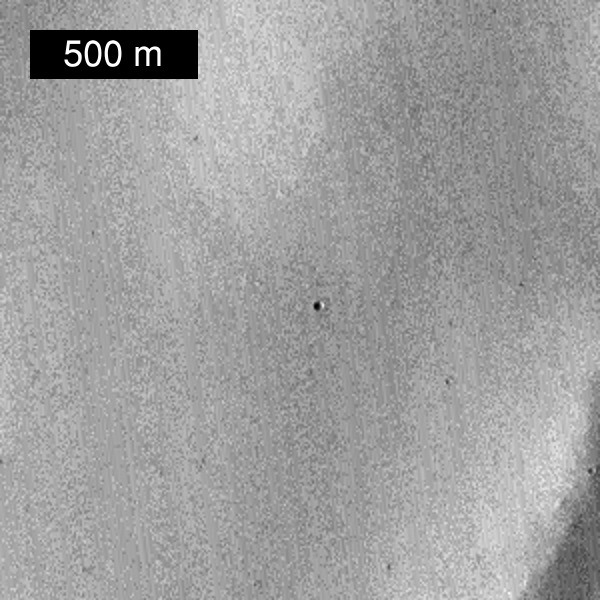}}
  \subfigure[February 29, 2016 (CTX {\scriptsize F23\_044972\_1721\_XN\_07S0980W})] 
  {\includegraphics[height=1.7in]{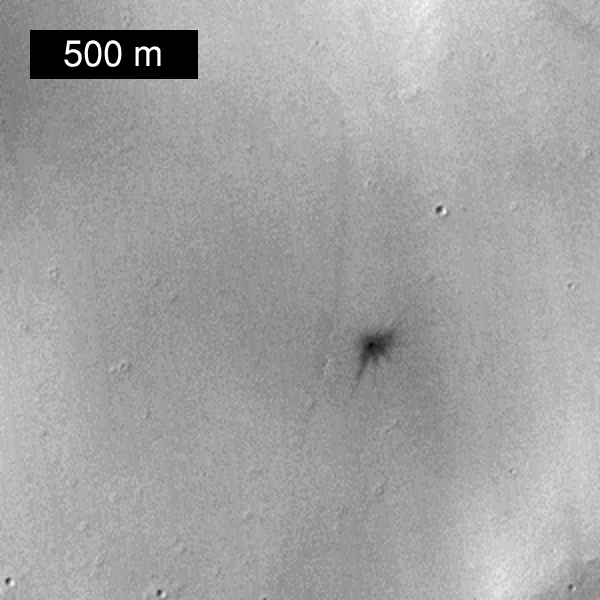}}
  \subfigure[October 3, 2020 (HiRISE ESP\_066506\_1720) at 8.07$^\circ$ S,
    261.5$^\circ$ E]
  {\includegraphics[height=1.7in]{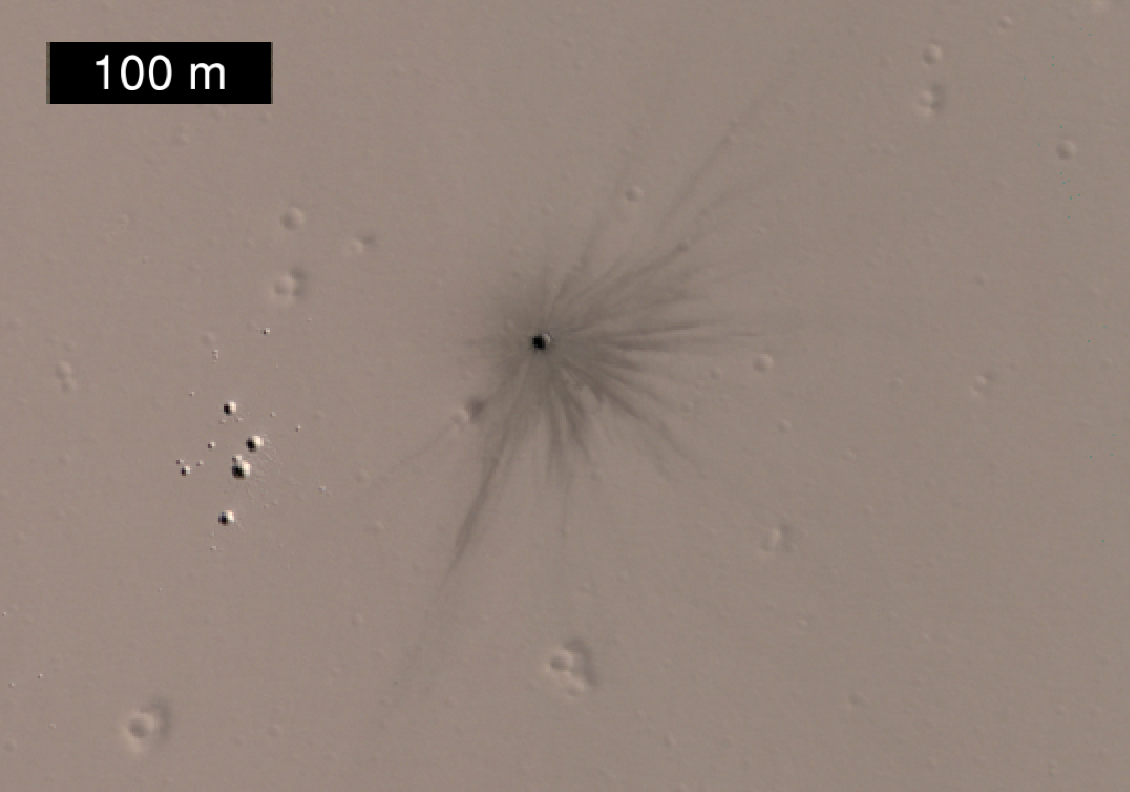}}
  \caption{Examples of fresh impacts discovered by the machine learning
    classifier in CTX images: (a,d) before impact, (b,e) after impact, (c,f)
    follow-up image by HiRISE. Image credits:
    NASA/JPL/MSSS (CTX) and NASA/JPL/University of Arizona (HiRISE).} 
  \label{fig:disc}
\end{figure}

\subsection{Review of candidates by thermal inertia}
\label{sec:method-binning}

\edit{In an attempt to reduce bias in machine learning candidates as a
function of thermal inertia, w}{W}e also performed a manual review \edit{of
candidates that were selected}{} using a stratified sampling
procedure\edit{}{ that partitioned candidates by their thermal inertia
values}. To perform the stratified sampling of candidates, we first 
computed a thermal inertia value for every candidate within the
detection database using the THEMIS thermal inertia map
(\unit[100]{m/pixel}) as a primary
source of measurements and using the TES map (\unit[7.5]{km/pixel}) as
a backup as described above for manually detected candidates.  In this
case, we used the value of thermal inertia pixel closest to the
center of the CTX window in which the candidate was found.
Then, a set of \num{1000}
candidates was obtained by selecting the top \num{100} candidates
\edit{}{(by confidence)} within
each of ten thermal inertia bins evenly spaced between \num{0} and
\num{1000} tiu. The review process followed the same procedure as
described above. 

\edit{
To estimate the statistical significance of biases in detection rates
across the various sets of candidates as a function of thermal
inertia, we developed a statistical test that supposes as a null
hypothesis that detection of a fresh impact is equally likely to occur
within each thermal inertia range on the surface. The fraction of
total detections within each thermal inertia bin is then compared with
the expected binomial null distribution of impacts that would occur
according to the fraction of the total surface contained within the
bin. Because this test is conducted for each bin, the Bonferroni
correction (Bonferroni, 1986) is applied to keep the
family-wise error rate across all tests below the $\alpha = 0.05$
significance level.
}{}

\section{Results and discussion}
\label{sect:results}

We first report on the top \num{1000} most-confident machine learning
candidates and characterize the new discoveries, then discuss
differences in candidates found by manual search versus machine
learning.  Finally, we report on an analysis by thermal inertia bins
that helps reduce the observational bias in the catalog.

\subsection{Highest-confidence machine learning candidates}

The manual review of the top 1000 machine learning candidates, ranked
in descending order of posterior probability, yielded \todo{169}
dateable fresh impacts.  Of these, \todo{99} were already recorded in
the catalog of known fresh impacts, providing confirmation of human
and machine agreement.  The remaining \todo{70} dateable fresh impacts
constitute the discovery of fresh impacts not previously known.
An additional set of \todo{465}
candidates \edit{appear to be}{visually resemble} fresh impacts, but the CTX archive does not
contain a ``before'' image that pre-dates the impact's occurrence, so
the date of formation cannot be constrained.  An additional \todo{166}
candidates were old or degraded impacts (not ``fresh''), and \todo{89}
candidates were not impact features (e.g., mistaken with shadows due
to terrain or other dark features on the surface). Because there is
some imprecision in determining the location of each candidate on the
surface, \todo{111} of the top 1000 candidates were duplicates with
other candidates and not automatically filtered using a distance-based
threshold of $\unit[600]{m}$. 

Two examples of the fresh impacts discovered by the machine
learning classifier are shown in Figure~\ref{fig:disc}.  The latest
CTX image pre-impact and the earliest CTX image post-impact are
accompanied by the follow-up image we requested from HiRISE.  The
first example was revealed to be a cluster of smaller impacts, while
the second one is a single impact with extensive rays.

\begin{figure}
  \centering
  \subfigure[\edit{All}{Manual} Detections]{\includegraphics[width=0.45\textwidth]{fig/ti_and_tone_v2.pdf}}
  \subfigure[Machine Learning Detections]{\includegraphics[width=0.45\textwidth]{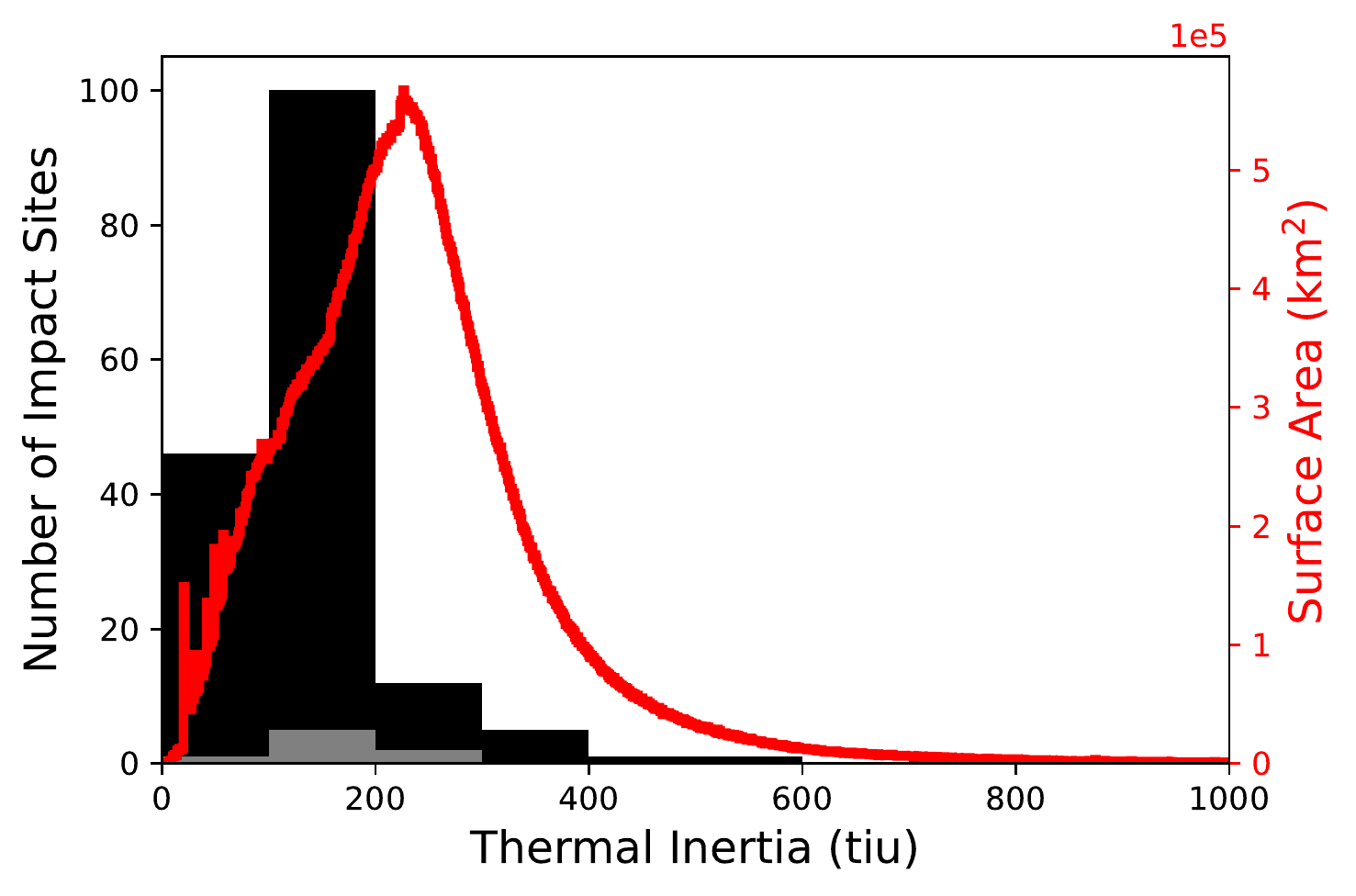}}
  \caption{Thermal inertia distribution for fresh impacts within -60
    to 60 degrees latitude compared with the overall thermal inertia
    distribution for Mars over the same area, derived from TES and
    THEMIS thermal inertia maps. Stacked histograms show a breakdown
    of detections by type.  The left panel shows results for \edit{the
    entire catalog}{manual detections}, replicated from Figure~\ref{fig:ti} for direct
    comparison with the ML detections only (right).  Both
    distributions show that the thermal inertia of fresh 
    impact sites is lower than that of Mars as a whole (peak near \num{150}
    tiu vs.~\num{225} tiu).} 
  \label{fig:ti-comp}
\end{figure}

\begin{figure}
  \centering
  \includegraphics[width=4in]{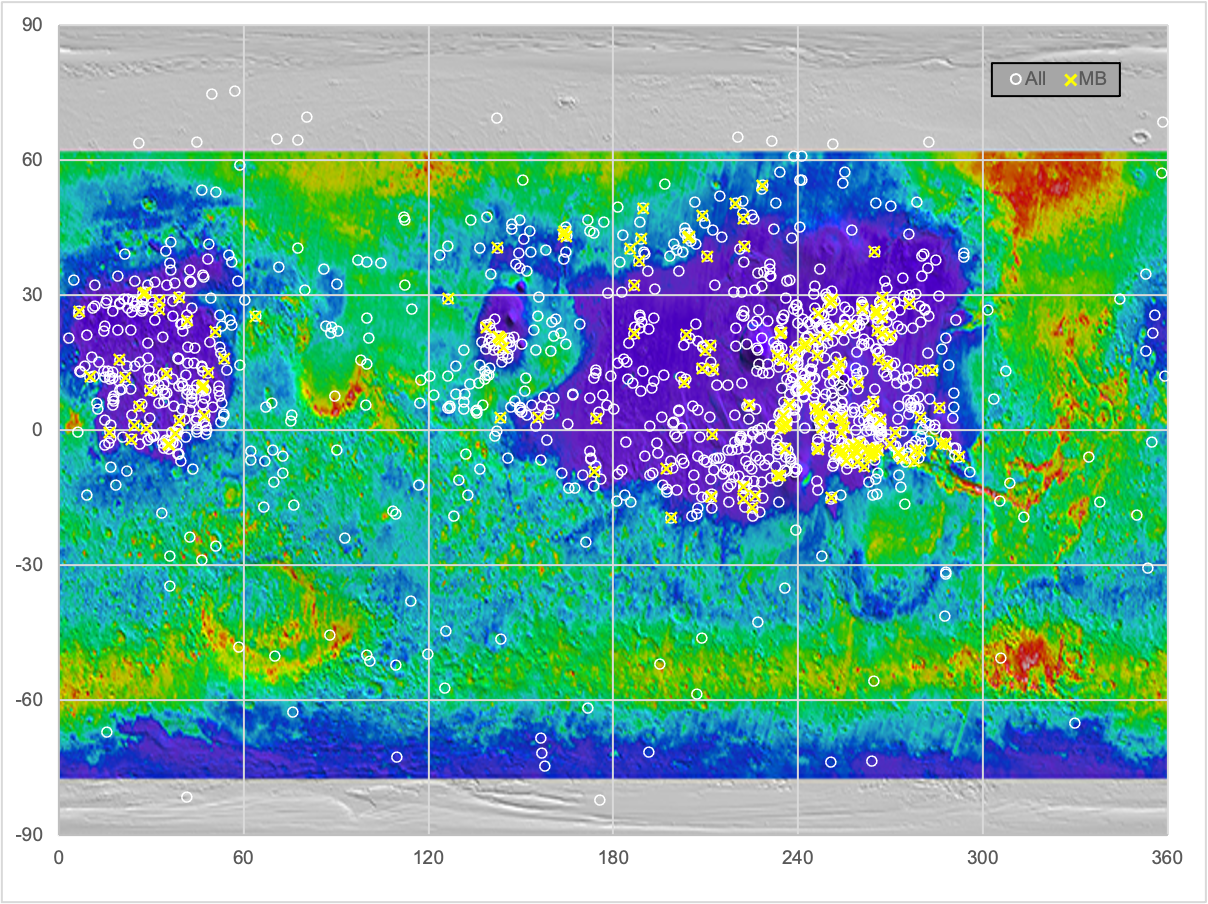}
  \caption{Spatial distribution of fresh impact sites on Mars overlaid
  on the TES thermal inertia basemap, augmented with ML-identified
  candidates (yellow x's).}
  \label{fig:cand-map}
\end{figure}

One of the primary questions driving this work was whether the machine
learning candidates would exhibit the same observational bias as those
identified by humans.  Indeed, the most-confident machine learning
candidates appear to replicate this bias and are concentrated
towards areas of low thermal inertia (see Figure~\ref{fig:ti-comp}).
This bias is also evident when the known impacts are plotted on a
map of Mars (Figure~\ref{fig:cand-map}).
This is not surprising, since the machine
learning classifier was trained on examples identified by humans; the
classifier is most confident about the same kinds of impacts that it was
trained on.  Its ability to exhaustively search in a consistent
fashion for the same kinds of features enabled the discovery of
new fresh impacts that are similar to those already in the catalog.
However, there are some interesting differences in the properties of
the impacts that were found via machine learning compared to the
pre-existing catalog.

\subsection{Differences in machine learning versus manually detected
  candidates}

\begin{figure}
  \centering
  \subfigure[June 27, 2016 (CTX {\scriptsize J04\_046501\_2223\_XI\_42N088W})]
  {\includegraphics[height=1.4in]{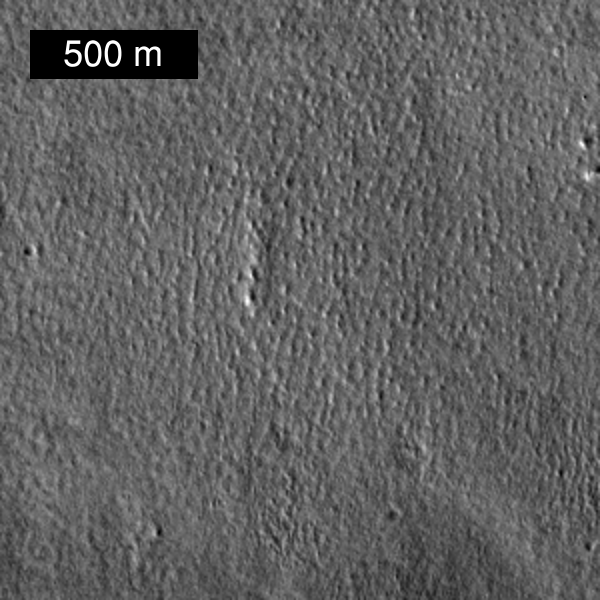}}
  \subfigure[May 9, 2017 (CTX {\scriptsize J15\_050549\_2220\_XI\_42N088W})]
  {\includegraphics[height=1.4in]{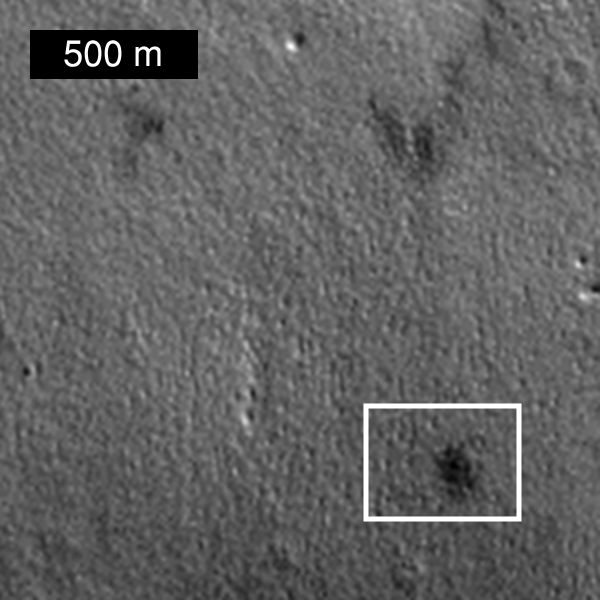}}
  \subfigure[April 6, 2019 (CTX {\scriptsize K16\_059503\_2220\_XI\_42N088W})]
  {\includegraphics[height=1.4in]{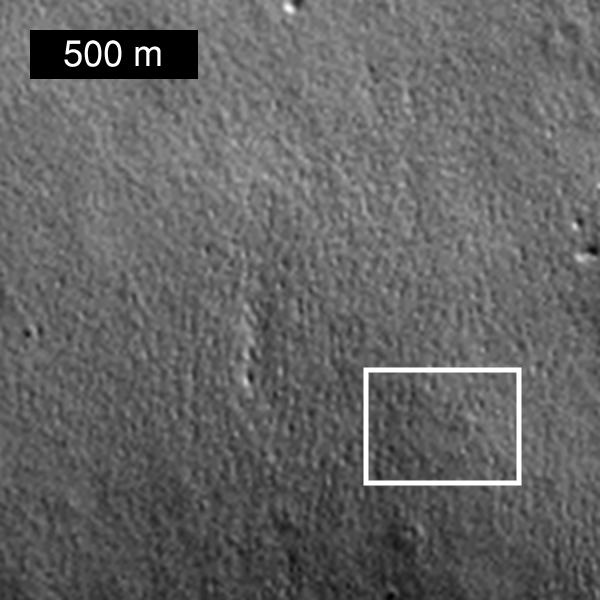}}
  \subfigure[August 26, 2020 (HiRISE ESP\_066822\_2225)]
  {\includegraphics[height=1.4in]{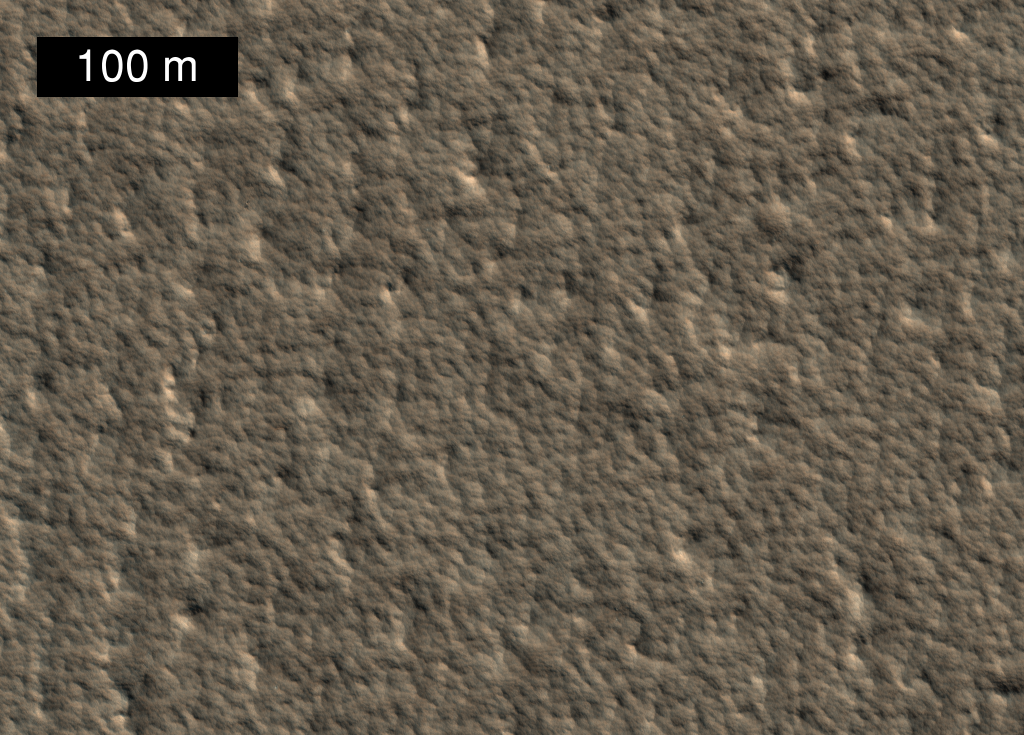}}
  \caption{Enigmatic machine learning candidate first observed on May
    9, 2017 (a vs.~b) that persisted through at least February 9, 2018
    and disappeared by April 6, 2019 (c). A HiRISE observation
    acquired on August 26, 2020 revealed no visible impact at the
    candidate location of 43.3$^\circ$ N, 271.8$^\circ$ E (d).}
  \label{fig:falsepos}
\end{figure}

Why were \todo{70} fresh impact sites detected via machine learning
but not by prior review of the CTX images?  If there were no
distinguishing features for these impacts, they could have been missed
by random chance or due to reviewer fatigue.  However, we found that there
are some factors that characterize this subpopulation and could help
explain how they were overlooked.  These factors point to 
advantages of employing semi-automated search techniques.

\begin{table}
  \caption{Location and properties of fresh impacts discovered via machine
    learning, sorted by thermal inertia.}
  \label{tab:impacts2}
  \begin{scriptsize}
  \begin{tabular}[9]{l|llllllll} \\ \hline
    Id & Latitude (N) & Longitude (E)
    & Type & Halo? & Rays? & \edit{D}{Effective d}iameter & Dust
    cover index & Thermal inertia \\ \hline
    \input{fig/ML-impacts-table2.tex}
    \hline
  \end{tabular}
  \end{scriptsize}
\end{table}

\begin{table}
  \caption{CTX before/after images and HiRISE confirmation images for
    fresh impacts discovered via machine learning, sorted by 
    thermal inertia.  Latitude is in degrees north and longitude is in
    degrees east.} 
  \label{tab:impacts}
  \begin{tiny}
  \begin{tabular}[8]{l|lllllll} \\ \hline
    Id & Latitude & Longitude & CTX before & Date & CTX after & Date 
    & HiRISE image \\ \hline
    \input{fig/ML-impacts-table1.tex}
    \hline
  \end{tabular}
  \end{tiny}
\end{table}

HiRISE follow-up high-resolution
images of the newly discovered impact sites were used to produce 
the measurements described in Section~\ref{sec:measure}.  An
examination of the HiRISE images revealed that one of the \todo{70}
new fresh impacts found in CTX images did not correspond to an impact
site.
Features identified in low-resolution imagery are sometimes
revealed to be transient or have other causes, which underscores the
value of following up with HiRISE for confirmation.  This
site is shown in Figure~\ref{fig:falsepos}.  The dark blotches
were first observed in May 2017 and persisted for at least nine Earth
months (February 2018), but they are not evident in CTX or HiRISE
imagery from April 
2019 onward.  We suspect this feature is the result of aeolian
redistribution of bright dust, but cannot be certain of that
explanation for this feature. 
%
As a result, we excluded this candidate and compiled measurements
for \todo{69} sites uniquely 
identified by the machine learning classifier and confirmed by manual
review.
\begin{wrapfigure}{r}{0.4\textwidth}
  \centering
    \includegraphics[width=2.5in]{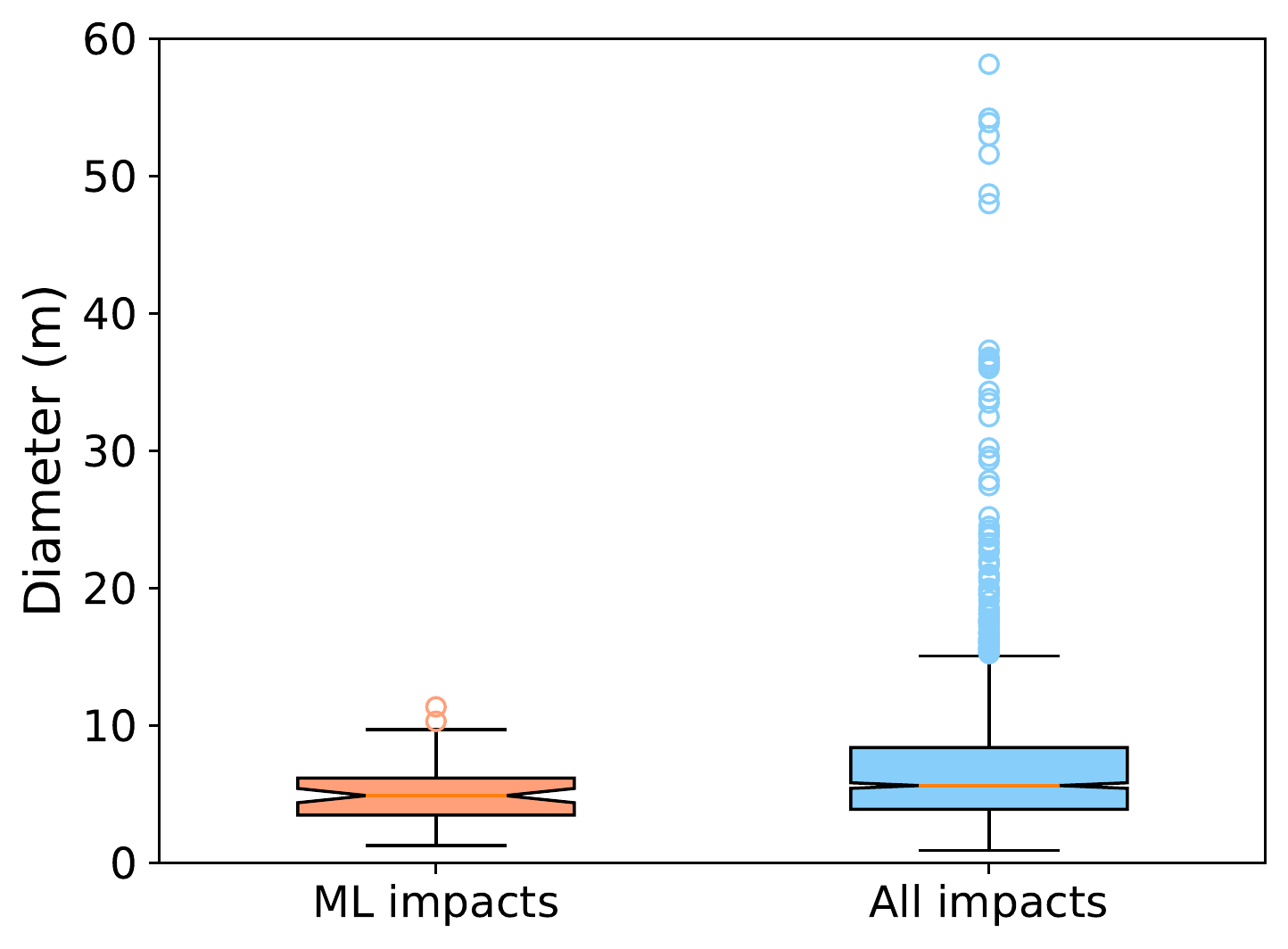}
  \caption{Distribution of fresh impact crater sizes (diameters) for
    impacts found uniquely by the machine learning classifier versus
    all impacts.  The box shows the inter-quartile range (IQR or span from
    25\% to 75\% of the distribution, whiskers extend the box by \num{1.5}
    times IQR, and outliers are shown with circles.}
  \label{fig:diam}
\end{wrapfigure}
%
We compiled information about each of the \todo{69} new discoveries,
including the location (latitude, longitude) and impact properties and
measurements (Table~\ref{tab:impacts2}) and
the before and after CTX images and HiRISE confirmation image
(Table~\ref{tab:impacts}).
Three of these sites were independently discovered by human
review during this study. We compared the characteristics of \todo{66}
ML impacts (uniquely found by the machine learning classifier) to
those of the entire catalog (including the ML impacts, \num{100}
additional impacts that were found by both ML and humans, and
\num{1058} previously identified impacts).

\paragraph{Impact size.} 
The diameters of the \todo{66} ML impact sites averaged $5.04 \pm 2.17$
m, while the full catalog averaged $7.22 \pm 6.09$ m.
The difference in distribution of diameter values is shown in
Figure~\ref{fig:diam}.  The full catalog has quite a few large
outliers that exceed \num{1.5} times the inter-quartile range.
This difference in distribution could indicate that ML is more
effective than humans at finding the smaller impact sites
\edit{}{\citep{delatte:mlcraters19,lagain:mars21}.} 
It could also partially  
explain the \edit{}{previously observed} difference in size-frequency
distribution slope between dated martian impacts
and \edit{the distribution seen at the Earth-Moon system}{those found
  on the Moon}. 
\edit{}{Mars has been characterized as having relatively fewer small new
  impacts than the Moon, but under-counting of these small impacts was
  identified as a possible reason~\citep{daubar:catalog22}.}
  We also find that ML detects clusters more
readily than they are found in the full catalog (\num{73}\% of ML sites are
clusters versus \num{59}\% of the full catalog).  \edit{These}{Clustered} impacts may
have \edit{more irregular spatial structure that are}{albedo patterns
  that are less symmetric and more spatially dispersed, which could
  be} easier for humans to overlook. 

\paragraph{Impact albedo features.}
The albedo features around new impact sites are what allow us to find
them in lower-resolution orbital images, so their nature is directly
connected to detectability and the resulting completeness of the
dataset. The most common features, halos (widespread diffuse areas of
differing 
albedo) are similarly common among both the full catalog (\num{88}\%) and
the ML subset (\num{86}\%). Rays (linear streaks emanating from the impact
site) also have similar rates of occurrence among both groups (\num{64}\% of
the full catalog vs.~\num{52}\% of the ML subset). The disturbed area around
the impact site could be darker or lighter than the surroundings, or
have some areas of dark and some areas of light. The ML sites have
similar distributions between these types as the full catalog, except
that dual-toned blast zones are less common among the ML sites (\num{8}\%
vs.~\num{14}\%). This could be explained by a scarcity of dual- and
light-toned impact sites in the training set, or the small numbers of
these types of sites in general. (Light-toned sites are 
so rare, only \num{4}\% of the full catalog, that finding none in this
subset is not surprising.) 

\paragraph{Surface characteristics.}
We also explored the surface characteristics that enable detection of
these new impacts to evaluate the effectiveness of ML in improving the
completeness of the data. The Dust Cover Index (DCI) around the ML
sites (average $0.9361 \pm 0.0074$) 
matches closely that of the catalog as a whole ($0.9377 \pm 0.0150$),
and thermal inertia of the surroundings were also similar ($142.0$
Jm$^{-2}$K$^{-1}$s$^{-0.5}$ for the ML sites vs.~$155.2$ for the whole
catalog). This demonstrates that the training set was effective at
training the classifier to find impacts on the same types of terrains.  

Overall, the primary distinguishing features of the ML sites were that
the impacts tended to be smaller and were more likely to be composed
of a cluster of impacts and therefore exhibited more variability in
their spatial appearance.  These impacts are likely more difficult
to find using solely human review of the CTX images.

\subsection{Fresh impacts and thermal inertia}

The preceding analysis focused on the most-confident candidates
identified by the machine learning classifier.  Since most of the
positive training examples came from areas of low thermal inertia, it
is unsurprising that the most confident candidates also come from low
thermal inertia areas.  Our expectation is that impacts are occurring
independently of the terrain type.  However, impacts in high thermal
inertia areas may be harder to visually detect, so if detected by the
classifier, they might not be evident in the set of most confidently
classified candidates.

\begin{figure}
  \centering
  \subfigure[November 11, 2016 (CTX {\scriptsize J09\_048129\_2079\_XN\_27N254W})]
  {\includegraphics[height=1.7in]{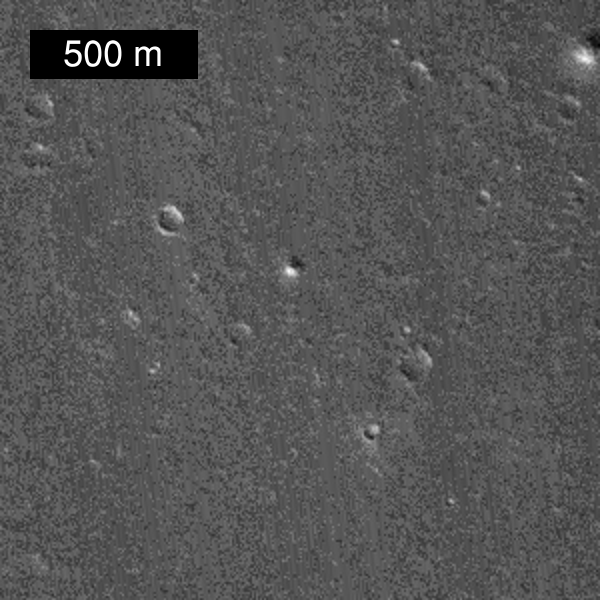}}
  \subfigure[May 4, 2019 (CTX {\scriptsize K17\_059852\_2080\_XN\_28N254W})]
  {\includegraphics[height=1.7in]{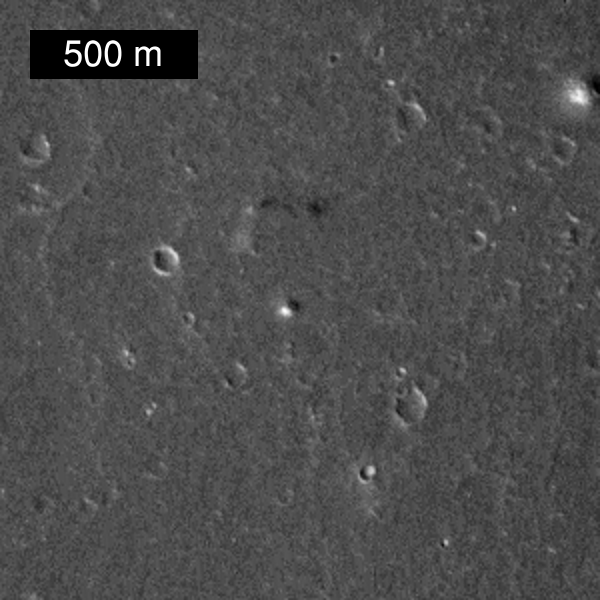}}
\caption{High thermal inertia fresh impact candidate discovered by
  the machine learning classifier at 27.5$^\circ$ N, 105.5$^\circ$ E.}
\label{fig:high-ti}
\end{figure}

\edit{As described in Section 2.4, w}{W}e \edit{}{therefore}
performed a deeper search \edit{}{, as described in
\autoref{sec:method-binning},}
  to determine whether the machine learning candidates stratified
by thermal inertia exhibit less bias.  We divided the fresh impacts
into bins based on thermal inertia and reviewed 100 candidates for
each bin.  This process yielded an additional five viable
candidates, for which we are in the process of obtaining HiRISE
follow-up images via HiWish.  An example of an impact with THEMIS
thermal inertia greater than \num{1000} tiu is shown in
Figure~\ref{fig:high-ti}.  The HiRISE follow-up image, when available,
will help us resolve its properties.

\begin{figure}
    \centering
    \subfigure[Manual Detections]{\includegraphics[width=0.3\textwidth]{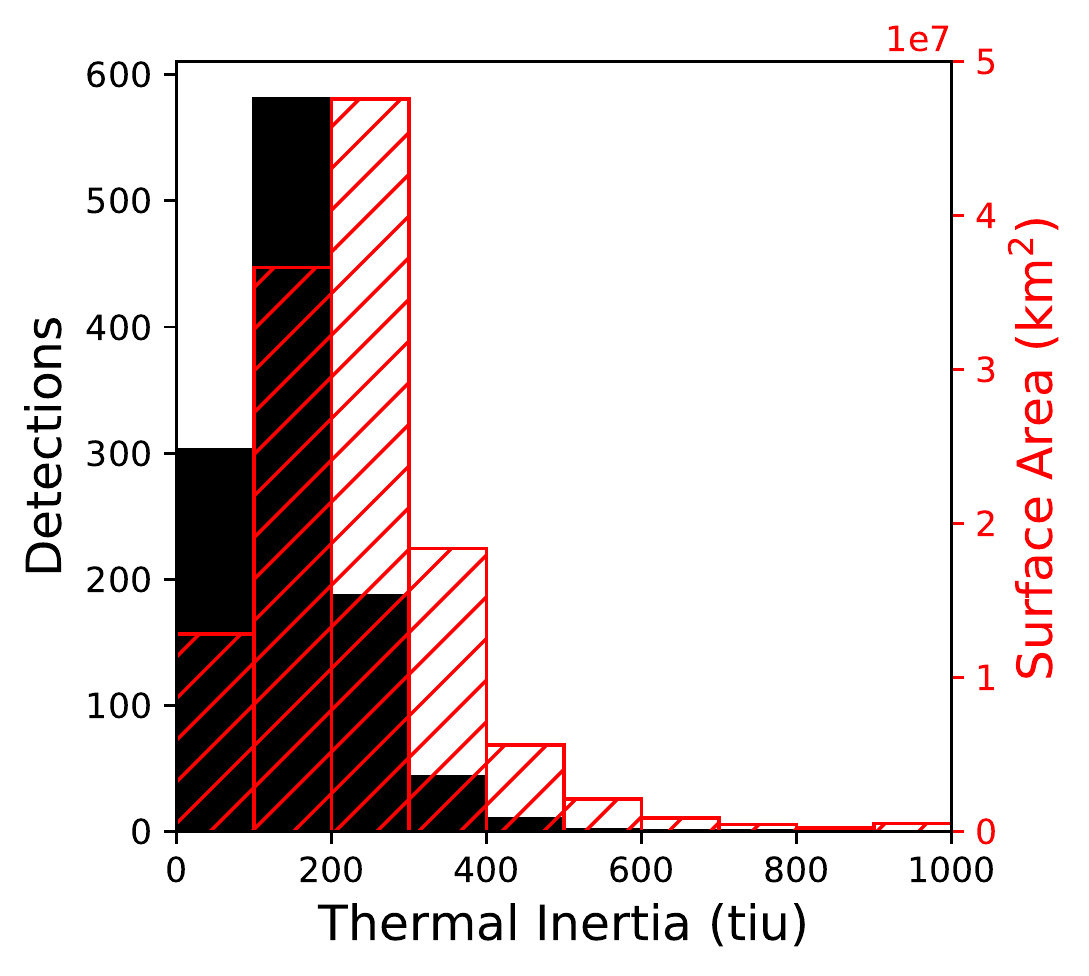}}
    \subfigure[Most-Confident ML Detections]{\includegraphics[width=0.3\textwidth]{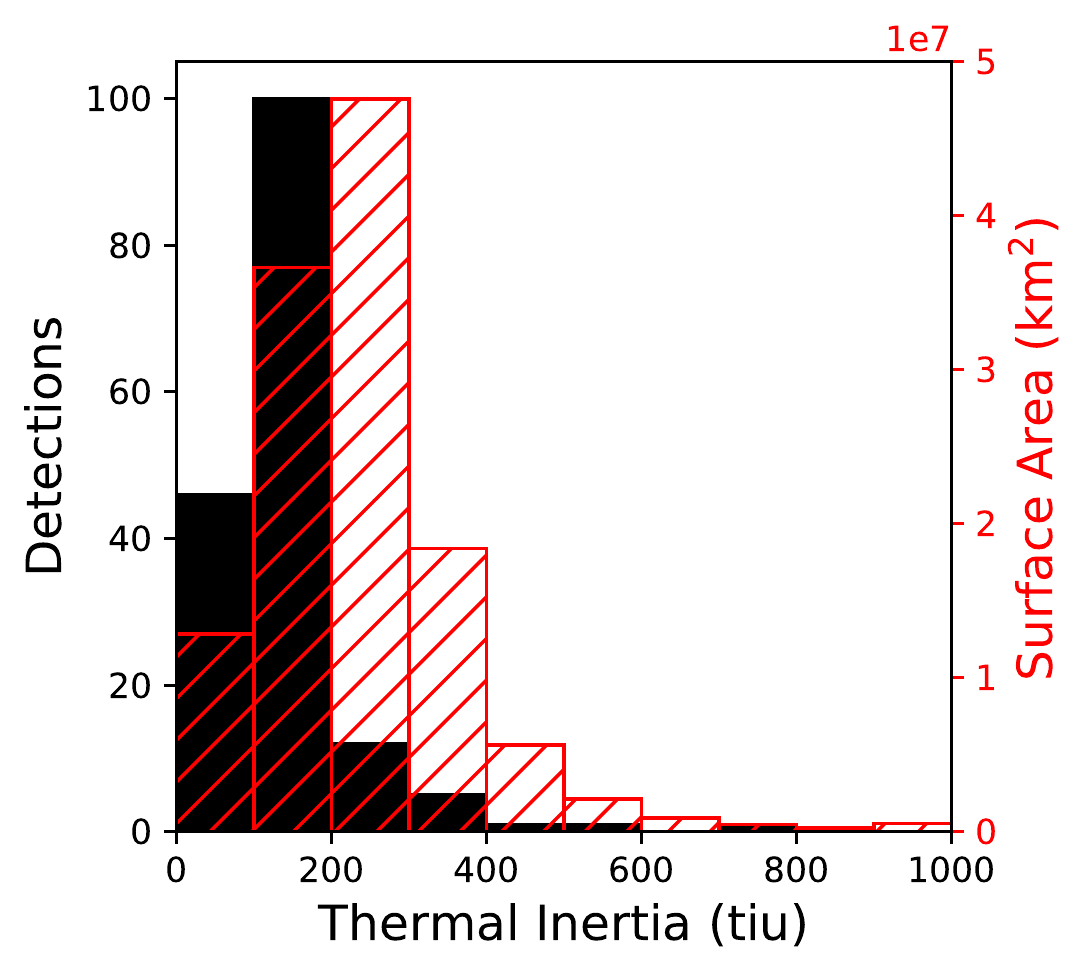}}
    \subfigure[Stratified ML Detections]{\includegraphics[width=0.3\textwidth]{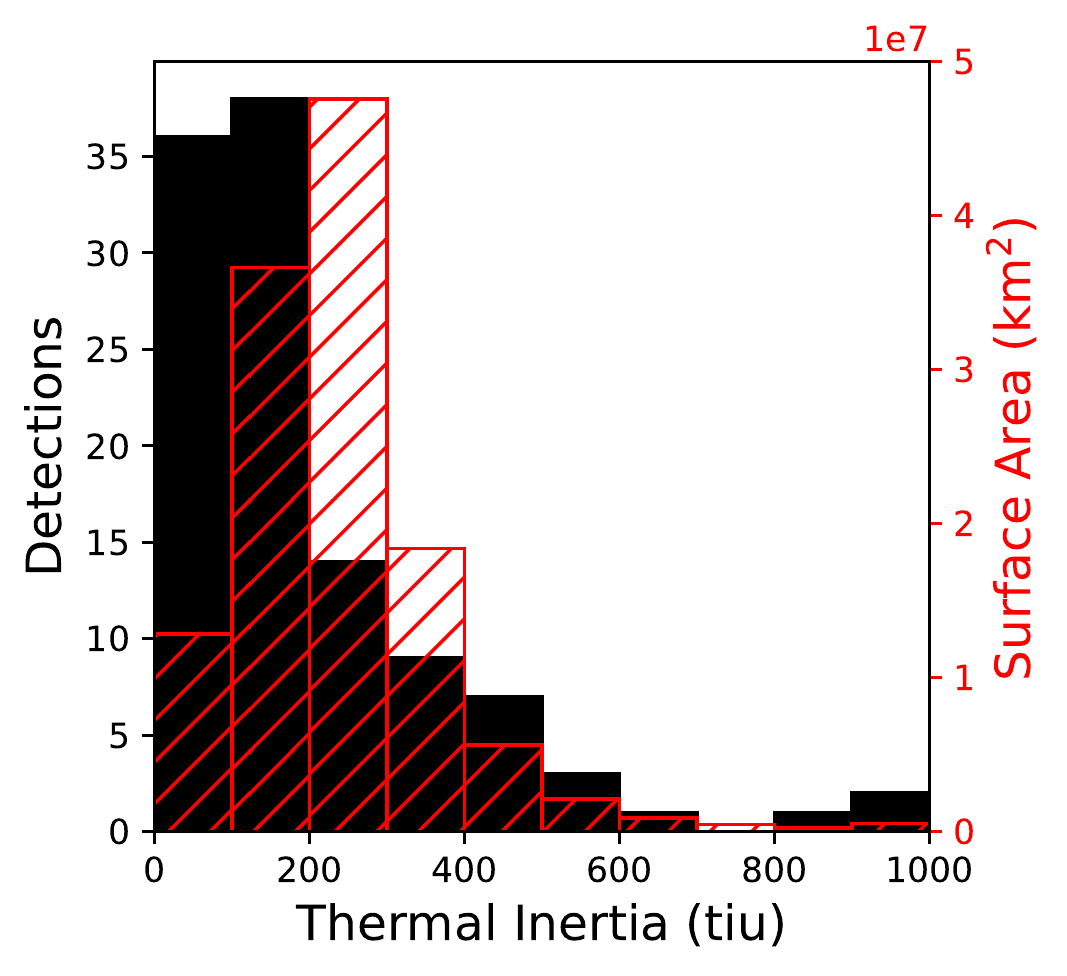}}
    \caption{Biases in new impact crater detection rates within 60
      degrees latitude from the equator for (a) human detections, (b)
      \edit{overall}{most-confident} ML detections, and (c)
      \edit{}{thermal-inertia} stratified ML 
      detections. \edit{Detection rates are computed by dividing the number
      of detections within each bin by the total surface area within
      the bin. Stars show when the rates for each bin differ
      significantly from the overall detection rate (dashed line), at
      an $\alpha = 0.05$ level, using a Bonferroni correction to
      account for multiple tests across bins. Detections above
      $\mathrm{tiu} = 500$ are combined to increase the sample size
      within the uppermost bin.}{Black bars show impact distribution
        by thermal inertia.  Red hashed bars show the counts expected
        given the Mars global thermal inertia distribution.}
        \edit{The results show that}{While the most-confident ML
          detections replicate the manual bias,} the
      stratification process \edit{helps to}{} reduce\edit{ the biases in ML
      detections}{s it}.} 
    \label{fig:individual-ti-ci-tests}
\end{figure}

\edit{}{\autoref{fig:individual-ti-ci-tests} shows the distribution of
impacts by their thermal inertia values (black bars) compared to the
counts that would be expected given the global distribution of
thermal inertia across Mars, assuming uniform observations across the
surface (red hashed bars).  The manually 
detected impacts exhibit a bias towards lower thermal inertia bins
that does not match the global distribution.  The most-confident
machine learning detections also exhibit this bias.  However,
stratifying the detections by thermal inertia bin allows the
detection of a larger proportion of impacts at higher thermal
inertia, providing a better fit to the global distribution.
Note that this strategy is possible only because the machine learning
classifier generated so many candidates across the entire archive.  
There does not exist a similarly voluminous sample of human detections
that can be stratified to reduce bias in a post-detection fashion.}

\edit{}{To quantify the observational bias, we measure the difference between
the observed ($O$) and expected ($E$) thermal inertia distribution of
impacts using the Kullback-Leibler (KL) divergence,
\begin{equation}
  D_{KL}(O \parallel E) = \sum_{i=1}^{10} O_i \log \frac{O_i}{E_i} ,
\end{equation}
where $O_i$ and $E_i$ refer to the probability of an impact occurring
in thermal inertia bin $i$.  The magnitude of $D_{KL}(O \parallel E)$
characterizes the bias present; if the two distributions were
identical, this value would be zero.  The $D_{KL}$ value for the
manual detections was \num{0.337}.  The most-confident ML
detections had a higher value of \num{0.528}, while stratifying by
thermal inertia corrected the bias to \num{0.316}, a value that is
lower than that of the manual detections.
}

\edit{}{Note that KL divergence naturally emphasizes differences where
there is high probability mass in the distribution.  In this case, it
focuses on differences in the lower \edit{}{thermal inertia
  (}TI\edit{}{)} bins where most of the 
detections occur.  The increase in detections in the highest TI bins
(tiu 500-1000) obtained by the stratified method, while important
scientifically, does not have as large of an influence on $D_{KL}$.
Also, while stratification reduces some bias, other biases could
remain in the data, such as biases based on the appearance of detected
impacts, such as the tone of the surrounding ejecta relative to the
background. A re-training of the model with additional relevant
examples would be required to mitigate these appearance-based
detection biases.}

\edit{To assess observational bias, we compared impact detection rates
within each thermal inertia bin to the overall 
detection rate. We also performed a similar comparison for the manual
detections and original list of unstratified machine learning
candidates, and the results are shown in Figure 10.

In each plot of Figure 10, the bars
indicate the conditional impact detection rates within each thermal
inertia bin. The detection rate is the total number of detections
divided by the surface area contained within each bin. The red dashed
line show the overall detection rate (i.e., the total number of
detections divided by the total surface area). Stars indicate bins for
which the rate differs significantly from the overall rate, using the
statistical test described in Section 2.4. Note that
due to differences in surface area covered by each bin, particularly
for the higher thermal inertia bins with low surface areas and low
numbers of detections, rates can have larger differences in
\emph{magnitude} from the overall rate but remain insignificant
compared with other lower thermal inertia bins that do have
significant differences. Accordingly, we combined the top five thermal
inertia bins to increase the sample size when determining whether
differences were significant in the uppermost TI range. 

Comparing the first two subplots that do not perform stratification,
we see similar patterns in rates across the bins, with many bins
differing significantly from the overall rate, indicating a strong
bias in detection rate as a function of thermal inertia. On the other
hand, the stratified machine learning candidates have more bins that
do not differ significantly from the overall rate, suggesting the
detection bias has been mitigated somewhat through this process. This
is consistent with the hypothesis that the underlying machine learning
classifier replicates the detection biases in the dataset on which it
was trained, whereas attempting to control for this bias through
stratification can mitigate some of these effects.
Note that this strategy is possible only because the machine learning
classifier generated so many candidates across the entire archive.  
There does not exist a similarly voluminous sample of human detections
that can be stratified to reduce bias in a post-detection fashion.}{}

\edit{
There are several caveats and limitations associated with
this analysis and the results. First is the assumption that a uniform
detection rate is expected across the surface of Mars, independent of
thermal inertia. Implicitly, this assumes (1) impacts are \textit{a
  priori} uniformly distributed across the surface and not dependent
on latitude or other geographic properties correlated with thermal
inertia, and (2) the surface has been evenly explored with CTX both
geographically and on average with similar intervals between the
before and after observations that can bound impact events. If these
assumptions approximately hold, then the remaining bias observed is
due solely to differences in detection rates. Otherwise, some amount
of bias is observation bias associated with the violations of
assumptions (1) or (2). Second, while stratification might reduce
biases in detection rates, other biases could remain in the data, such
as biases based on the appearance of detected impacts, such as the
tone of the surrounding ejecta relative to the background. A
re-training of the model would be required to mitigate these
appearance-based detection biases.}{} 

\section{Conclusions}
\label{sect:conc}

We trained a convolutional neural network on thousands of positive and
negative examples of fresh impacts and deployed it to search
\num{112207} images previously obtained 
by the Mars Reconnaissance Orbiter (MRO)'s Context Camera (CTX).  After
classifying more than 2 billion CTX image windows of size \num{1.8}
$\times$ \num{1.8} km, the top \num{1000} candidate detections (by classifier
confidence) were manually reviewed.  This process yielded \num{69} fresh
impacts that had not been previously identified in the data.
Most of the newly discovered impacts were also
in low-TI regions, demonstrating consistency with the examples the
model was trained on, which are primarily also in low-TI areas.

Each of the fresh impacts that were identified by the machine learning
classifier were further investigated using HiRISE on MRO.
The high-resolution image enabled the measurement of
key impact properties such as size, multiplicity, the presence of
rays, etc.
During this follow-up process,
three candidates were independently found by human review.
We found that the properties of the \todo{66} impacts uniquely
identified by
the classifier were largely consistent with those of the full catalog.
However, the ML-discovered impacts were more likely to be
clusters of multiple impacts, and they tended to be smaller than those
in the full catalog.  These impacts are therefore likely more
heterogeneous and somewhat harder to detect with manual review,
highlighting a useful role for the classifier in increasing coverage.

\edit{}{The bias towards areas of low thermal inertia in the catalog
  of known impacts can be addressed by assessing candidates across the
  TI range, not just the most confident ones.}
We \edit{also}{} partitioned the classifier candidates based on the thermal
inertia of the terrain surrounding the candidate and reviewed the top
\num{100} candidates for each of \num{10} thermal inertia ranges.
This enabled the identification of five additional impacts that
had not been previously found by manual detection.  HiWish requests
for follow-up imaging of these impacts have also been made and will
enable future measurements of their properties.  In addition, this
analysis yielded four new discoveries of fresh impacts on
terrain with thermal inertial of at least \num{400} tiu,
a subpopulation that is rare within the catalog.
\edit{}{This analysis was possible only because the classifier
  generated so many candidates across the globe, with more than one
  million candidates that had a posterior probability of at least
  \num{0.99}.} 

We conclude that machine learning is a useful technique for
increasing the detection of fresh impacts on Mars.  The use of an
automated classifier can accelerate the rate at which fresh impacts
are found and catalogued.  Time is still required to review the
machine learning candidates, but this time is focused on the areas
most likely to yield new discoveries.  The classifier provides a
useful pointer to focus attention where it is most likely to pay off.




\comment{
\begin{figure}[thb]
	\centering
		\includegraphics[<options>]{}
	  \caption{}\label{fig1}
\end{figure}

\begin{table}[<options>]
\caption{}\label{tbl1}
\begin{tabular*}{\tblwidth}{@{}LL@{}}
\toprule
  &  \\ 
\midrule
 & \\
 & \\
 & \\
 & \\
\bottomrule
\end{tabular*}
\end{table}



}


\section*{Acknowledgments}

We thank Marko Green and Justin Martia for assisting with the labeling
of the CTX data set, the CTX and PDS teams for providing the data,
Steven Lu and Masha Liukis for developing the distributed machine
learning system used to train the classifier, and the Jet Propulsion
Laboratory R\&TD program for funding this work.
The High Performance Computing resources used in this investigation
were provided by funding from the JPL Information and Technology
Solutions Directorate.
Part of this research was carried out at the Jet Propulsion
Laboratory, California Institute of Technology, under a contract with
the National Aeronautics and Space Administration.  \copyright \: \num{2021}.
All rights reserved.

AG, DW, JP and IJD were partially supported by NASA grant
80NSSC20K0789.
VTB was supported by a fellowship within the IFI programme of the
German Academic Exchange Service (DAAD).

We are grateful to the HiRISE operations team for acquiring our HiWish
target suggestions in a timely manner. 

\printcredits

\bibliographystyle{cas-model2-names}

\bibliography{impacts}

\bio{}
\endbio

\endbio

\end{document}

%% file: fig/ML-impacts-table2.tex
     1	& 9.1990 & 242.6790 & cluster & y & y & 7.41 & 0.9369 & 44.91\\
     2	& 39.6142 & 264.8912 & cluster & y & n & 6.29 & 0.9218 & 45.14\\
     3	& -6.1622 & 274.1629 & cluster & y & n & 2.64 & 0.9348 & 60.66\\
     4	& -3.0650 & 287.9550 & single & n & n & 4.50 & 0.9448 & 63.82\\
     5	& -14.5268 & 225.9985 & cluster & y & y & 7.88 & 0.9382 & 74.78\\
     6	& 29.5490 & 39.0180 & single & y & n & 3.26 & 0.9260 & 75.56\\
     7	& 30.3690 & 27.3720 & cluster & y & n & 3.17 & 0.9254 & 79.27\\
     8	& 23.3190 & 257.0400 & cluster & y & y & 4.97 & 0.9260 & 80.80\\
     9	& -4.8700 & 253.9290 & cluster & y & n & 9.06 & 0.9384 & 83.49\\
    10	& 11.8900 & 10.0560 & single & y & y & 2.88 & 0.9322 & 84.97\\
    11	& -5.8643 & 292.3527 & cluster & y & n & 3.33 & 0.9441 & 87.10\\
    12	& 1.7600 & 254.4290 & single & y & y & 5.67 & 0.9315 & 88.62\\
    13	& 0.6796 & 255.0087 & cluster & y & n & 5.85 & 0.9317 & 90.05\\
    14	& 8.9010 & 29.6060 & cluster & y & y & 4.34 & 0.9385 & 91.43\\
    15	& -4.3845 & 279.6718 & single & y & n & 2.75 & 0.9446 & 92.94\\
    16	& -4.2270 & 256.5710 & cluster & y & n & 9.72 & 0.9349 & 96.07\\
    17	& 22.55271 & 253.8839 & cluster & y & y & 5.66 & 0.9354 & 99.41\\
    18	& -0.4373 & 16.4684 & cluster & y & n & 4.64 & 0.9467 & 100.37\\
    19	& -2.859 & 259.977 & cluster & n & y & 9.17 & 0.9406 & 100.67\\
    20	& 12.7930 & 48.4590 & single & y & y & 3.83 & 0.9346 & 101.01\\
    21	& -4.4959 & 264.2976 & cluster & y & y & 4.99 & 0.9372 & 102.31\\
    22	& -5.8110 & 263.4880 & cluster & n & n & 7.08 & 0.9484 & 104.50\\
    23	& 29.0800 & 251.0380 & cluster & n & y & 4.81 & 0.9274 & 108.49\\
    24	& -1.9970 & 23.5660 & single & y & y & 4.11 & 0.9456 & 109.37\\
    25	& 21.3150 & 249.4010 & cluster & y & y & 5.31 & 0.9319 & 110.57\\
    26	& 2.6430 & 174.4250 & single & y & n & 1.26 & 0.9312 & 114.37\\
    27	& 2.8903 & 254.1615 & cluster & y & y & 11.36 & 0.9301 & 116.43\\
    28	& 16.6122 & 246.3148 & cluster & n & y & 10.31 & 0.9342 & 125.21\\
    29	& 40.4860 & 142.3931 & single & y & n & 4.90 & 0.9405 & 126.44\\
    30	& -3.9690 & 235.9680 & single & y & y & 5.00 & 0.9358 & 129.60\\
    31	& 43.5112 & 163.9189 & single & n & n & 2.30 & 0.9435 & 129.78\\
    32	& 12.762 & 251.467 & single & n & y & 3.63 & 0.9379 & 131.16\\
    33	& -0.4776 & 271.6420 & cluster & y & y & 4.05 & 0.9327 & 131.25\\
    34	& 2.9980 & 247.4600 & single & y & n & 4.98 & 0.9428 & 131.59\\
    35	& -4.1960 & 262.5030 & cluster & y & y & 5.67 & 0.9389 & 132.86\\
    36	& -8.0650 & 261.4580 & single & y & y & 2.37 & 0.9432 & 133.25\\
    37	& 20.8956 & 142.5618 & cluster & y & n & 6.64 & 0.9356 & 133.86\\
    38	& -14.8476 & 250.9032 & cluster & n & n & 3.74 & 0.9552 & 136.46\\
    39	& -4.6036 & 264.0991 & cluster & y & y & 5.75 & 0.9364 & 137.27\\
    40	& 0.3060 & 28.4810 & cluster & y & n & 4.42 & 0.9473 & 139.08\\
    41	& -2.9244 & 270.0695 & single & y & n & 2.25 & 0.9423 & 140.54\\
    42	& 20.6012 & 142.8866 & single & y & y & 3.70 & 0.9278 & 142.74\\
    43	& 21.5430 & 267.2350 & cluster & y & n & 7.25 & 0.9283 & 143.87\\
    44	& -4.9936 & 259.1517 & cluster & y & y & 5.19 & 0.9433 & 148.18\\
    45	& -0.6070 & 254.0170 & cluster & n & n & 8.51 & 0.9341 & 148.48\\
    46	& 2.7460 & 279.8160 & single & y & y & 2.84 & 0.9409 & 148.97\\
    47	& -6.8390 & 277.7210 & cluster & y & n & 5.38 & 0.9480 & 150.71\\
    48	& -4.4870 & 258.6750 & cluster & y & y & 6.34 & 0.9319 & 151.02\\
    49	& 47.6020 & 209.0360 & cluster & y & n & 4.73 & 0.9337 & 151.63\\
    50	& -6.7113 & 255.7250 & cluster & n & n & 9.07 & 0.9519 & 154.55\\
    51	& 27.0131 & 260.7333 & cluster & y & n & 4.91 & 0.9262 & 161.00\\
    52	& 26.8140 & 32.5150 & cluster & y & y & 1.94 & 0.9237 & 163.99\\
    53	& -4.4250 & 254.3330 & cluster & y & n & 7.15 & 0.9353 & 164.55\\
    54	& 2.8290 & 155.6070 & single & y & n & 1.50 & 0.9319 & 167.59\\
    55	& 13.20978 & 279.55647 & cluster & y & y & 3.76 & 0.9454 & 167.65\\
    56	& -0.9353 & 212.0407 & cluster & y & n & 6.28 & 0.9369 & 175.90\\
    57	& 25.9570 & 246.6694 & cluster & y & y & 6.81 & 0.9320 & 176.46\\
    58	& 18.8686 & 211.5511 & cluster & y & n & 4.90 & 0.9305 & 177.76\\
    59	& 10.6180 & 259.5470 & cluster & y & y & 5.64 & 0.9384 & 183.64\\
    60	& 11.7909 & 21.4820 & cluster & n & y & 4.08 & 0.9346 & 201.62\\
    61	& -10.0940 & 233.1460 & cluster & y & y & 4.37 & 0.9335 & 213.04\\
    62	& 24.2920 & 41.5100 & cluster & y & y & 4.11 & 0.9309 & 215.12\\
    63	& 15.84881 & 53.525 & cluster & y & n & 2.45 & 0.9362 & 217.94\\
    64	& -3.2060 & 35.4300 & cluster & y & y & 3.27 & 0.9440 & 227.34\\
    65	& 13.646 & 252.922 & single & n & n & 3.67 & 0.9367 & 228.22\\
    66	& 28.4210 & 250.4310 & cluster & y & y & 5.55 & 0.9190 & 315.92\\
    67	& 21.8120 & 50.8230 & single & y & n & 2.27 & 0.9374 & 324.99\\
    68	& 26.3710 & 6.5540 & cluster & y & y & 3.41 & 0.9263 & 346.44\\
    69	& 25.8970 & 265.3760 & cluster & y & y & 7.96 & 0.9325 & 394.27\\

%% file: fig/ML-impacts-table1.tex
     1	& 9.1990 & 242.6790 & P18\_008064\_1868\_XN\_06N117W & 2008-04-15 & J04\_046344\_1906\_XI\_10N117W & 2016-06-15 & ESP\_066533\_1895\\
     2	& 39.6142 & 264.8912 & B20\_017280\_2195\_XN\_39N095W & 2010-04-03 & F03\_036822\_2191\_XN\_39N095W & 2014-06-04 & ESP\_067007\_2200\\
     3	& -6.1622 & 274.1629 & P17\_007496\_1738\_XI\_06S086W & 2008-03-02 & G14\_023873\_1727\_XN\_07S085W & 2011-08-30 & ESP\_072283\_1740\\
     4	& -3.0650 & 287.9550 & F04\_037428\_1779\_XN\_02S072W & 2014-07-21 & J02\_045459\_1770\_XN\_03S071W & 2016-04-07 & ESP\_070252\_1770\\
     5	& -14.5268 & 225.9985 & F03\_036916\_1674\_XI\_12S134W & 2014-06-12 & K02\_054032\_1674\_XN\_12S134W & 2018-02-04 & ESP\_068315\_1655\\
     6	& 29.5490 & 39.0180 & B20\_017407\_2108\_XN\_30N321W & 2010-04-13 & J03\_045969\_2118\_XN\_31N321W & 2016-05-17 & ESP\_066857\_2100\\
     7	& 30.3690 & 27.3720 & B07\_012423\_2099\_XN\_29N332W & 2009-03-21 & F03\_036923\_2096\_XN\_29N332W & 2014-06-12 & ESP\_067781\_2105\\
     8	& 23.3190 & 257.0400 & P22\_009607\_1715\_XI\_08S127W & 2008-08-14 & K12\_057896\_1705\_XN\_09S127W & 2018-12-02 & ESP\_066915\_2035\\
     9	& -4.8700 & 253.9290 & B18\_016740\_1742\_XN\_05S106W & 2010-02-20 & J12\_049561\_1737\_XI\_06S105W & 2017-02-21 & ESP\_067377\_1750\\
    10	& 11.8900 & 10.0560 & G07\_020797\_1918\_XI\_11N349W & 2011-01-03 & J07\_047592\_1917\_XN\_11N350W & 2016-09-20 & ESP\_067122\_1920\\
    11	& -5.8643 & 292.3527 & D02\_028158\_1732\_XI\_06S067W & 2012-07-29 & D20\_034896\_1724\_XI\_07S067W & 2014-01-05 & ESP\_069764\_1740\\
    12	& 1.7600 & 254.4290 & G04\_019972\_1991\_XI\_19N148W & 2010-10-30 & K04\_054995\_1990\_XN\_19N148W & 2018-04-20 & ESP\_067720\_1820\\
    13	& 0.6796 & 255.0087 & P16\_007233\_1808\_XN\_00N104W & 2008-02-11 & K11\_057816\_1799\_XN\_00S105W & 2018-11-26 & ESP\_070029\_1805\\
    14	& 8.9010 & 29.6060 & P17\_007584\_1915\_XI\_11N330W & 2008-03-09 & D16\_033442\_1915\_XI\_11N330W & 2013-09-14 & ESP\_067306\_1890\\
    15	& -4.3845 & 279.6718 & P17\_007654\_1730\_XN\_07S080W & 2008-03-14 & J04\_046501\_1730\_XN\_07S080W & 2016-06-27 & ESP\_068907\_1755\\
    16	& -4.2270 & 256.5710 & F05\_037706\_1761\_XI\_03S103W & 2014-08-12 & K10\_057196\_1739\_XN\_06S103W & 2018-10-09 & ESP\_067179\_1755\\
    17	& 22.55271 & 253.8839 & D18\_034238\_2026\_XN\_22N106W & 2013-11-15 & K17\_059873\_2046\_XI\_24N106W & 2019-05-05 & ESP\_069607\_2030\\
    18	& -0.4373 & 16.4684 & B02\_010261\_1786\_XI\_01S343W & 2008-10-04 & F23\_044704\_1797\_XN\_00S343W & 2016-02-08 & ESP\_066792\_1795\\
    19	& -2.859 & 259.977 & P20\_008683\_1754\_XN\_04S099W & 2008-06-03 & J19\_052132\_1754\_XI\_04S099W & 2017-09-09 & ESP\_065464\_1770\\
    20	& 12.7930 & 48.4590 & G14\_023802\_1932\_XN\_13N311W & 2011-08-25 & D15\_033072\_1950\_XN\_15N311W & 2013-08-16 & ESP\_067556\_1930\\
    21	& -4.4959 & 264.2976 & P17\_007615\_1764\_XN\_03S096W & 2008-03-11 & B18\_016555\_1759\_XN\_04S095W & 2010-02-06 & ESP\_068406\_1755\\
    22	& -5.8110 & 263.4880 & F19\_043073\_1745\_XN\_05S096W & 2015-10-04 & K05\_055402\_1716\_XI\_08S096W & 2018-05-22 & ESP\_067469\_1740\\
    23	& 29.0800 & 251.0380 & P04\_002763\_2080\_XN\_28N109W & 2007-02-27 & P11\_005453\_2080\_XN\_28N108W & 2007-09-25 & ESP\_067245\_2095\\
    24	& -1.9970 & 23.5660 & B02\_010419\_1771\_XI\_02S336W & 2008-10-16 & K06\_055569\_1756\_XI\_04S336W & 2018-06-04 & ESP\_067491\_1780\\
    25	& 21.3150 & 249.4010 & P04\_002473\_2019\_XN\_21N110W & 2007-02-05 & J01\_045368\_2032\_XN\_23N110W & 2016-03-31 & ESP\_067034\_2015\\
    26	& 2.6430 & 174.4250 & D02\_028004\_1840\_XN\_04N185W & 2012-07-17 & F20\_043630\_1858\_XN\_05N185W & 2015-11-17 & ESP\_066773\_1825\\
    27	& 2.8903 & 254.1615 & P18\_008235\_1828\_XI\_02N106W & 2008-04-29 & K12\_058093\_1807\_XN\_00N105W & 2018-12-18 & ESP\_066902\_1830\\
    28	& 16.6122 & 246.3148 & P03\_002183\_1972\_XI\_17N113W & 2007-01-13 & P12\_005717\_1973\_XI\_17N113W & 2007-10-15 & ESP\_066480\_1970\\
    29	& 40.4860 & 142.3931 & G22\_026739\_2235\_XN\_43N218W & 2012-04-10 & D20\_034954\_2216\_XN\_41N217W & 2014-01-10 & ESP\_069334\_2210\\
    30	& -3.9690 & 235.9680 & G22\_026683\_1749\_XI\_05S124W & 2012-04-05 & K05\_055337\_1784\_XN\_01S124W & 2018-05-17 & ESP\_069278\_1760\\
    31	& 43.5112 & 163.9189 & D07\_029758\_2236\_XN\_43N195W & 2012-12-01 & D15\_033239\_2214\_XI\_41N195W & 2013-08-29 & ESP\_069320\_2240\\
    32	& 12.762 & 251.467 & P20\_008881\_1913\_XN\_11N108W & 2008-06-18 & F05\_037627\_1917\_XN\_11N108W & 2014-08-06 & ESP\_065517\_1930\\
    33	& -0.4776 & 271.6420 & G04\_019693\_1786\_XN\_01S088W & 2010-10-08 & K05\_055283\_1786\_XN\_01S088W & 2018-05-13 & ESP\_067627\_1795\\
    34	& 2.9980 & 247.4600 & P05\_002816\_1820\_XI\_02N112W & 2007-03-03 & D02\_028120\_1842\_XI\_04N112W & 2012-07-26 & ESP\_066612\_1830\\
    35	& -4.1960 & 262.5030 & G21\_026537\_1784\_XN\_01S097W & 2012-03-25 & K22\_061996\_1784\_XI\_01S097W & 2019-10-18 & ESP\_067126\_1760\\
    36	& -8.0650 & 261.4580 & P15\_006890\_1717\_XN\_08S098W & 2008-01-15 & F23\_044972\_1721\_XN\_07S098W & 2016-02-29 & ESP\_066506\_1720\\
    37	& 20.8956 & 142.5618 & P11\_005457\_2016\_XI\_21N217W & 2007-09-25 & K13\_058387\_2015\_XI\_21N217W & 2019-01-10 & ESP\_069466\_2010\\
    38	& -14.8476 & 250.9032 & P17\_007576\_1672\_XI\_12S109W & 2008-03-08 & B03\_010714\_1652\_XI\_14S109W & 2008-11-08 & ESP\_069172\_1650\\
    39	& -4.6036 & 264.0991 & P17\_007615\_1764\_XN\_03S096W & 2008-03-01 & F21\_043996\_1763\_XI\_03S096W & 2015-12-15 & ESP\_068010\_1755\\
    40	& 0.3060 & 28.4810 & B05\_011566\_1800\_XI\_00N331W & 2009-01-13 & K11\_057666\_1787\_XI\_01S331W & 2018-11-14 & ESP\_067438\_1805\\
    41	& -2.9244 & 270.0695 & B09\_013034\_1786\_XN\_01S090W & 2009-05-08 & K12\_058132\_1766\_XN\_03S090W & 2018-12-21 & ESP\_068287\_1770\\
    42	& 20.6012 & 142.8866 & G05\_020291\_2008\_XN\_20N217W & 2010-11-24 & G13\_023126\_2009\_XN\_20N217W & 2011-07-03 & ESP\_070389\_2010\\
    43	& 21.5430 & 267.2350 & F16\_041886\_2016\_XI\_21N092W & 2015-07-04 & K15\_059134\_2016\_XI\_21N092W & 2019-03-09 & ESP\_066598\_2020\\
    44	& -4.9936 & 259.1517 & B02\_010384\_1754\_XI\_04S101W & 2008-10-13 & J10\_048585\_1729\_XI\_07S100W & 2016-12-07 & ESP\_067997\_1750\\
    45	& -0.6070 & 254.0170 & P07\_003752\_1796\_XN\_00S105W & 2007-05-15 & P21\_009237\_1800\_XI\_00N106W & 2008-07-16 & ESP\_067311\_1795\\
    46	& 2.7460 & 279.8160 & P08\_004173\_1812\_XI\_01N080W & 2007-06-17 & J05\_046791\_1808\_XI\_00N079W & 2016-07-20 & ESP\_067310\_1830\\
    47	& -6.8390 & 277.7210 & B02\_010568\_1726\_XN\_07S082W & 2008-10-27 & D04\_028897\_1729\_XN\_07S082W & 2012-09-25 & ESP\_028897\_1730\\
    48	& -4.4870 & 258.6750 & G09\_021856\_1765\_XN\_03S101W & 2011-03-26 & J10\_048875\_1730\_XI\_07S101W & 2016-12-29 & ESP\_067390\_1755\\
    49	& 47.6020 & 209.0360 & P20\_008711\_2286\_XN\_48N151W & 2008-06-05 & D22\_035624\_2280\_XN\_48N151W & 2014-03-03 & ESP\_066943\_2280\\
    50	& -6.7113 & 255.7250 & B01\_010160\_1757\_XI\_04S104W & 2008-09-26 & J21\_052831\_1730\_XI\_07S104W & 2017-11-03 & ESP\_070385\_1730\\
    51	& 27.0131 & 260.7333 & P17\_007760\_2069\_XN\_26N099W & 2008-03-23 & G21\_026471\_2073\_XN\_27N099W & 2012-03-20 & ESP\_069910\_2075\\
    52	& 26.8140 & 32.5150 & G21\_026453\_2074\_XN\_27N327W & 2012-03-18 & K23\_062558\_2073\_XI\_27N327W & 2019-12-01 & ESP\_067319\_2070\\
    53	& -4.4250 & 254.3330 & G09\_021632\_1756\_XN\_04S105W & 2011-03-09 & F06\_038049\_1754\_XN\_04S105W & 2014-09-08 & ESP\_066968\_1755\\
    54	& 2.8290 & 155.6070 & P15\_006762\_1831\_XI\_03N204W & 2008-01-05 & D18\_034123\_1816\_XN\_01N204W & 2013-11-06 & ESP\_067341\_1830\\
    55	& 13.20978 & 279.55647 & B17\_016317\_1920\_XN\_12N080W & 2010-01-18 & J05\_046580\_1938\_XN\_13N080W & 2016-07-04 & ESP\_069738\_1935\\
    56	& -0.9353 & 212.0407 & P14\_006694\_1776\_XN\_02S147W & 2007-12-31 & K04\_055061\_1773\_XN\_02S147W & 2018-04-25 & ESP\_069081\_1790\\
    57	& 25.9570 & 246.6694 & P02\_001761\_2054\_XN\_25N113W & 2006-12-11 & G05\_020274\_2069\_XN\_26N113W & 2010-11-23 & ESP\_067641\_2060\\
    58	& 18.8686 & 211.5511 & G04\_019972\_1991\_XI\_19N148W & 2010-10-30 & K04\_054995\_1990\_XN\_19N148W & 2018-04-20 & ESP\_068355\_1990\\
    59	& 10.6180 & 259.5470 & B18\_016832\_1895\_XI\_09N100W & 2010-02-28 & J10\_048809\_1925\_XI\_12N100W & 2016-12-24 & ESP\_067192\_1905\\
    60	& 11.7909 & 21.4820 & P08\_004222\_1917\_XI\_11N338W & 2007-06-21 & F02\_036541\_1943\_XI\_14N338W & 2014-05-13 & ESP\_068824\_1920\\
    61	& -10.0940 & 233.1460 & P22\_009607\_1715\_XI\_08S127W & 2008-08-14 & K12\_057896\_1705\_XN\_09S127W & 2018-12-02 & ESP\_066916\_1700\\
    62	& 24.2920 & 41.5100 & F05\_037661\_2046\_XN\_24N318W & 2014-08-09 & J15\_050452\_2046\_XI\_24N318W & 2017-05-01 & ESP\_067543\_2045\\
    63	& 15.84881 & 53.525 & P11\_005368\_1967\_XN\_16N306W & 2007-09-18 & K03\_054592\_1963\_XN\_16N306W & 2018-03-20 & ESP\_069588\_1960\\
    64	& -3.2060 & 35.4300 & G22\_026664\_1765\_XN\_03S324W & 2012-04-04 & J18\_051982\_1764\_XI\_03S324W & 2017-08-28 & ESP\_066633\_1770\\
    65	& 13.646 & 252.922 & B19\_016885\_1951\_XI\_15N107W & 2010-03-04 & K04\_055099\_1913\_XN\_11N106W & 2018-04-28 & ESP\_065385\_1940\\
    66	& 28.4210 & 250.4310 & P07\_003752\_2085\_XN\_28N109W & 2007-05-15 & F17\_042414\_2075\_XI\_27N109W & 2015-08-14 & ESP\_066559\_2085\\
    67	& 21.8120 & 50.8230 & F02\_036764\_2018\_XN\_21N309W & 2014-05-31 & K04\_055080\_2045\_XN\_24N309W & 2018-04-27 & ESP\_066883\_2020\\
    68	& 26.3710 & 6.5540 & B08\_012938\_2091\_XN\_29N353W & 2009-04-30 & J22\_053552\_2064\_XI\_26N353W & 2017-12-29 & ESP\_066502\_2065\\
    69	& 25.8970 & 265.3760 & P13\_006270\_2064\_XN\_26N094W & 2007-11-28 & J15\_050457\_2070\_XI\_27N094W & 2017-05-02 & ESP\_067073\_2060\\